# Zero-shot adaptable task planning for autonomous construction robots: a comparative study of lightweight single and multi-AI agent systems

Hossein Naderi[1], Alireza Shojaei[2], Lifu Huang[3], Philip Agee[4], Kereshmeh Afsari[5], Abiola Akanmu[6]

[1]Dept. of Building Construction, Myers-Lawson School of Construction, Virginia Tech, Blacksburg, VA. ORCID: https://orcid.org/0000-0002-6625-1326. Email: hnaderi@vt.edu
[2]Assistant Professor, Dept. of Building Construction, Myers-Lawson School of Construction, Virginia Tech, Blacksburg, VA (corresponding author). ORCID: https://orcid.org/0000-0003-3970-0541. Email: shojaei@vt.edu
[3]Assistant Professor, Computer Science Department, UC Davis, Davis, CA. Email: lfuhuang@ucdavis.edu
[4]Assistant Professor, Dept. of Building Construction, Myers-Lawson School of Construction, Virginia Tech, Blacksburg, VA, email: pragee@vt.edu
[5]Assistant Professor, Dept. of Construction Engineering and Management, Myers-Lawson School of Construction, Virginia Tech, Blacksburg, VA, email: keresh@vt.edu
[6]Associate Professor, Dept. of Construction Engineering and Management, Myers-Lawson School of Construction, Virginia Tech, Blacksburg, VA, email: abiola@vt.edu

**Abstract**

Robots are expected to play a major role in the future construction industry but face challenges due to high costs and difficulty adapting to dynamic tasks. This study explores the potential of foundation models to enhance the adaptability and generalizability of task planning in construction robots. Four models are proposed and implemented using lightweight, open-source large language models (LLMs) and vision language models (VLMs). These models include one single agent and three multi-agent teams that collaborate to create robot action plans. The models are evaluated across three construction roles: Painter, Safety Inspector, and Floor Tiling. Results show that the four-agent team outperforms the state-of-the-art GPT-4o in most metrics while being ten times more cost-effective. Additionally, teams with three and four agents demonstrate the improved generalizability. By discussing how agent behaviors influence outputs, this study enhances the understanding of AI teams and supports future research in diverse unstructured environments beyond construction.

## 1. Introduction

The construction industry is facing persistent challenges, including productivity improvement [1], [2], a severe workforce shortage [3], and safety issues [4]. On one hand, while other industries have enjoyed productivity gains during recent decades, the construction industry is struggling with stagnant productivity in both management and labor level [2], [5], [6]. This challenge not only extends the duration of critical infrastructure projects but also increases costs and reduces overall efficiency [7]. On the other hand, the industry faces a severe labor shortage, reaching 650,000 unfilled positions by 2024, which has slowed the completion of projects across both residential and infrastructure sectors [8]. Moreover, construction sites account for more than 20% of job-related fatalities in the US private sector, highlighting the hazardous nature of the work environment [9].

Different studies emphasized that these industry-wide problems cannot be addressed by marginal improvements to traditional construction methods, which have reached their technical limits [10], [11], [12]. Among multiple options, robotics in construction has held one of the greatest potential solutions for these challenges [13], [14]. Autonomous construction robots offer the potential to handle repetitive tasks, improving productivity and helping to enable construction in settings not currently feasible, e.g., for use in disaster relief, extra-terrestrial construction, or other dangerous and challenging environments [15]. However, when it comes to robotic adoption in construction sites, there are several challenges that impede them from achieving their full potential, including technological, economic, organizational, and economic aspects [16], [17].

Within technological challenges, several studies [16], [17], [18], [19], [20] have identified the lack of versatility, adaptability, and generalizability as major challenges preventing current construction robotic technologies from reaching the desired level of autonomy. One major category of these technologies is pre-programmed robots that can automatically perform single tasks with high precision (e.g. "Jailbot" [21] for drilling or "SAM" [22] for bricklaying or BIM-based robots [23], [24]). However, these robots are deterministically programmed for a specific construction task, thus even small variations in tasks or environmental conditions require significant reprogramming. Another category is data-driven paradigms, including machine-learning [25], reinforcement learning [26], and imitation learning [27], which struggle with similar adaptability and generalizability issues. These approaches rely on narrowly defined task-specific datasets, which restrict their effectiveness in construction environments that are often unpredictable and characterized by unique situations. When encountering a scenario that was not included in their training data, these approaches typically require retraining and dataset adjustment to adapt. Moreover, construction companies view their data as a valuable asset and are often reluctant to share it for model training purposes [28], [29], which further complicates efforts to improve adaptability with these data-driven approaches.

One solution to address adaptability challenges is to integrate human cognitive capabilities with robots through tele-operations paradigms. However, these approaches also fall short of achieving desired autonomy, as their dependency on human operators as well as the potential lag in sending and receiving data limits their performance and productivity. For example, a recent study [30] demonstrated that even slight connection delays can significantly impede robot performance in remote locations. Therefore, further studies are needed to explore solutions to improve adaptability in autonomous construction robots.

While adaptability can be applied to a wide range of functions in construction robotics, task planning serves as a foundational element that enables autonomous robots to function adaptively in dynamic environments [31]. Task planning involves high-level reasoning to determine which actions to take and in what sequence, based on the specific situation at hand. This process is the first step in bridging abstract goals with actionable tasks, making it essential for bringing task-level autonomy to robotics [32], [33]. For example, a robot assigned to paint a wall might plan the following steps: pick up a roller, dip it into the paint can, and apply the paint to the wall. This high-level planning defines the sequence of actions required to complete the task. However, it is not cost-effective for construction companies, especially small companies such as specialty contractors, to invest in robots that can only perform a certain type of work and remain inactive during other phases of a project [19]. Instead, adaptable and multi-task robots capable of performing various tasks are needed. In the absence of adaptable task planning, when the robot is needed for another task, such as floor tiling, pre-programmed and conventional data-driven approaches often fail because they require significant reprogramming or retraining to address the unpredictable events which are inherent in construction sites. By contrast, an adaptable and generalizable planning system would allow the robot to flexibly interpret the new task, identify the tools and steps required (e.g., laying adhesive, placing tiles, and ensuring alignment), and execute the task successfully. Such adaptability mirrors human problem-solving and underscores the need for improving the adaptability and generalizability of task planning systems for construction robots.

Recently, pre-trained Foundation Models, such as Large Language Models (LLMs), have been employed as Artificial Intelligence (AI) agents to demonstrate some level of adaptability and generalizability [34], [35]. Unlike traditional AI systems trained on task-specific datasets, these models are trained on massive, internet-scale datasets [36]. This training approach, combined with the Transformer architecture [37], enabled them to show commonsense understanding and find zero-shot solutions to downstream tasks that were not explicitly present in their training data [38]. These qualities of generalizability and adaptability can be seen as a potential solution to address the adaptability challenges in task planning for construction robotics.

However, there are significant limitations to Foundation Model applications in this context. On one hand, state-of-the-art API-based (closed-source) models, such as GPT-4o [39], are expensive to use as task-planning tools for construction robots. On the other hand, open-source powerful models, such as Llama-70b [40], require high computational power that is rarely available on most construction robots. Additionally, recent studies [41] indicate that lightweight open-source models often fall short in reasoning capabilities, limiting their effectiveness for task planning in complex environments like construction sites. Emerging research suggests that deploying multiple LLMs in cooperative systems can improve performance in complex cognitive and reasoning tasks by breaking problems into smaller components and engaging in debate or collaboration [42], [43], [44]. Therefore, further research is needed to explore the potential of both single and multi-LLM-powered agent systems as a solution to enhance generalizability and adaptability in automated task planning for autonomous construction robots.

The overarching research question for this study is: How can foundation models improve generalizability in task planning for autonomous construction robots across various construction tasks? To address this question, this study focuses on three primary objectives. First, it aims to propose and implement a single-agent system along with three multi-agent systems, i.e., two-agents, three-agents, and four-agents, for task planning of autonomous construction robots across three different roles: painter tradesperson, safety inspector, and floor tiling tradesperson (reasoning for this selection is detailed in Section 4.1). Second, it seeks to compare and analyze the performance of these systems together and against GPT4o, as the state-of-the-art commercial LLM model, based on five evaluation metrics, including correctness, temporal understanding, executability, time, and cost. Finally, the paper analyzes the achieved generalizability and adaptability achieved by these and discusses the trade-offs between single-agent and various multi-AI agent systems for a better understanding of key considerations for future studies.

This paper contributes to the field in several areas. First, it provides a comparative analysis of how different agent architectures, ranging from single-agent to multi-agent systems, affect the performance of the task planning on robots executing three different constriction activities. Second, this study demonstrates an improved generalizability across three defined roles by proposed three- and four-agent systems. Additionally, by evaluating detailed performance metrics, the study offers valuable insights into optimizing design and implementation for future multi-agent systems. Furthermore, this study can serve as a benchmark that can be utilized by other researchers to test various agent architectures and tasks, particularly in unstructured environments such as construction sites. This approach helps future researchers and practitioners to track the progress toward a fully automated and resilient future construction industry.

Section 2 provides the necessary background to define the scope of this paper and highlights how it distinguishes itself from related studies. Section 3 presents the proposed framework for building multi-AI agents and discusses the theoretical foundations that guide the design choices. In Section 4, the configuration and experimental procedures are detailed. Section 5 offers an overview of the implementation and the outputs of the models. The results obtained from the experiments are demonstrated and compared in Section 6. Section 7 analyzes the findings to enhance understanding, and identify the study limitations, and areas for future research (section 8). Finally, the paper is concluded in Section 9.

## 2. Background
### 2.1. Task planning

Task planning in robotics refers to the process of creating sequences of actions for robots to achieve specific goals autonomously [45]. These plans are typically generated based on models that describe the environment and the robot's capabilities. Several methods have been developed to enable robot task planning [46]. Classical planning frameworks such as Stanford Research Institute Problem Solver [47] and Planning Domain Definition Language [48] define tasks by specifying preconditions and effects for each action, allowing robots to plan by sequencing actions from the initial state to the goal. These classical methods assume a fully observable and deterministic environment, which makes them effective for structured problems but limited in real-world construction applications due to their limitation in handling incomplete information [49].

More advanced methods such as temporal planning and probabilistic planning are employed to address classical deterministic challenges [50]. Temporal planning [48] considers the duration and timing of actions, essential for coordinating activities in dynamic environments. Probabilistic planning [51], on the other hand, incorporates uncertainty by modeling environments with unpredictable outcomes, using techniques such as Markov Decision Processes [52] and Partially Observable Markov Decision Processes [53]. More recent advancements have explored the integration of data-driven techniques, including reinforcement learning and imitation learning, to learn from data and enhance task planning capabilities in complex situations [54]. However, these data-driven paradigms suffer from training on narrow and specific-task datasets [55], hindering their applications in real-world situations where robots face lots of objects and elements that are not available in the dataset. For example, a construction robot tasked with sorting debris after demolition may encounter a wide variety of unexpected materials not represented in its training dataset, resulting in ineffective task planning. This lack of adaptability and generalizability underscores the need for further research to develop more generalizable task planning methods capable of addressing the unpredictable and diverse challenges construction robots face.

### 2.2. Foundation model-powered AI agents

AI agents, built upon foundation models [36] such as Large Language Models (LLMs) and Vision-Language Models (VLMs), are revolutionizing various domains with their capabilities to find solutions for downstream tasks that are not included in their training dataset (zero-shot solutions) [56]. Unlike previous specialized AI models that were trained on task-specific data to perform certain tasks, these very large neural networks are trained on a broad and massive scale using internet-scale unlabeled data in an unsupervised manner. This training approach makes them adaptable to a wide range of general downstream tasks [38]. Recently, AI agents, such as GPT-4 [39], have been integrated for various robotic tasks, such as manipulation [57], [58], planning [59], Human-Robot Interaction (HRI) [60], [61], [62]. These advancements have the potential to pave the way for even more sophisticated applications, particularly adopting robots in unstructured environments of construction sites [63].

Building on advances of single agents and inspired by collaboration in human teams for solving complex tasks, many studies are focusing on multi-AI agent systems to address more complex challenges [64], [65]. In this approach, LLM-powered agents take on multiple roles and tasks, collaborating to achieve common goals similar to human teams. Recent studies suggest that such multi-agent systems can enhance factuality and reasoning [66], planning [67], and divergent thinking [68]. Building on these findings, this study hypothesizes that the integration of multi-agent AI systems can improve adaptability of task planning in construction robotics.

## 2.3. Autonomy vs Automation

To clearly define the scope of this paper, it is important to distinguish between the concepts of autonomy and automation, which are sometimes used interchangeably. In the context of construction robotics, automation, and autonomy represent two distinct levels of robotic functionality and decision-making (or cognitive) capability [69]. Automation refers to the execution of predefined tasks with limited human intervention [70]. Automated systems follow specific instructions or rules to perform repetitive or structured activities, such as material handling or precise assembly tasks [71]. These systems are typically limited to the tasks they are explicitly programmed to perform and lack the ability to adapt to variations or unforeseen circumstances without human input [70], [72].

Conversely, autonomy encompasses a higher level of decision-making (cognitive) and generalizability (adaptability) [73], [74]. Autonomous robots have the capability to perceive their environment, make informed decisions, and execute actions independently to achieve specific goals [69]. Autonomy involves complex cognitive processes such as task planning, problem-solving, and real-time adaptation to dynamic and unstructured environments commonly found on construction sites. Autonomy enables robots to handle diverse tasks and unpredictable scenarios, enhancing their effectiveness in complex construction tasks [73].

Various taxonomies and classification systems have been developed to conceptualize autonomy. For example, the Society of Automotive Engineers (SAE) [75] introduced a scale for autonomous vehicles, ranging from Level 0 (manual control) to Level 5 (fully autonomous). In most frameworks, a level of autonomy signifies a function that was previously carried out by a human. However, these discrete classification systems suggest an "all-or-nothing" approach to robotic autonomy, which does not fully capture the multi-disciplinary nature of autonomy [76]. Instead, autonomy is best understood as existing along a continuum, ranging from low to high across different functions, reflecting the varied and flexible roles autonomy can play in different contexts [76]. This study adapted the robot taxonomy developed by Parasuraman et al. [77], which provided a continuum range for the main components and tasks of a robot: including sense, plan, and act (see Figure 1). According to this, each of the" sense, plan, and act" primitives can be allocated to either the human or the robot (or both), thus, a robot can vary in autonomy level (from low to high) along the sense, plan, and act. The focus of this study is on Plan primitive as discussed in more detail in section 2.5.

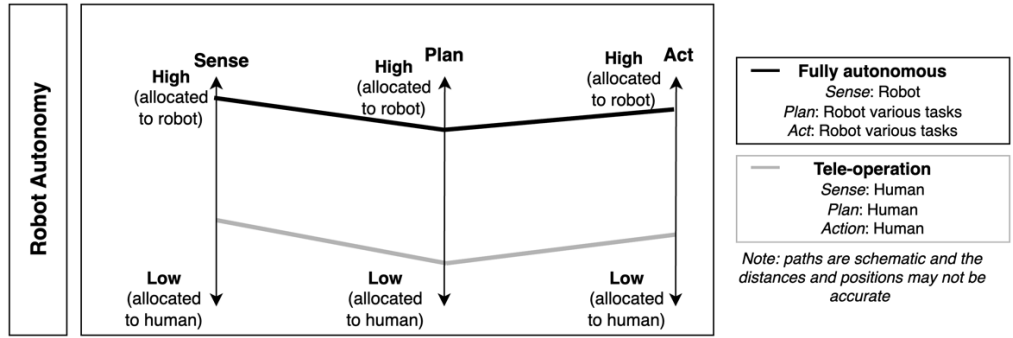

*Figure 1* Robot Autonomy based on Sense, Plan, Act model (adapted from [77]).

### 2.4. Generalizability

Generalizability (or adaptability and versatility) has been highlighted in several sections above as a missing characteristic in existing construction robots. This section defines the concept and examines its relationship with autonomy. In the robotic context, generalizability refers to the ability of a robot or agent to apply learned skills or behaviors to new, previously unseen tasks [78]. This ability is critical as it allows robots to be more adaptable, flexible, and efficient in various situations without requiring explicit reprogramming or retraining for each unique task. This definition closely aligns with related terms such as versatility, flexibility, and adaptability in construction settings, which are frequently highlighted as critical yet often missing characteristics in existing construction robotics technologies [16], [17], [18]. This capability is crucial given the inherently dynamic and unpredictable nature of construction sites, where conditions can vary significantly from one project to another and even within the same project over time.

Generalizability and autonomy are intrinsically connected in the context of robotics, with generalizability serving as an inherent characteristic of systems exhibiting high autonomy. This relationship is evident in various definitions of autonomy. For example, Thrun [79] defined autonomous systems as those with the ability to accommodate variations in their environment. Similarly, Murphy [80] described autonomy as the capability of a robot to adapt to changes in its environment. These definitions highlight that autonomy is not simply about performing a task independently; it is fundamentally about responding to dynamic conditions and unforeseen challenges.

### 2.5. Point of Departure

This section distinguishes the focus of this paper from other related studies. Figure 2 highlights the plan primitive of robot autonomy, which is the main focus of this paper, among the three primitives depicted in Figure 1. Additionally, it roughly positions this paper within the plan autonomy range to provide clarity on its scope and contribution. In most existing studies, robots receive detailed, written commands as input that specify the task's intention and the objects required to use, implying a need for human intervention to fully interpret and plan the task toward its goals. For example, in ProgPrompt [81] paper, robots were instructed with commands like

"bring coffeepot and cupcake to the coffee table.". This clearly demonstrates that the robot requires human intervention to identify relevant objects and understand the task's goal. Similarly, the SayCan framework [82] provided the input, "I spilled my drink, can you help?", indicating that the robot relied on human-provided sentences to understand the task's intention. Another example is Huang et al. [83], who investigated the use of LLMs as zero-shot planners by providing commands such as "Get the glass of milk" to robots. These examples collectively demonstrate that current studies, despite improving the generalizability, are still dependent on human intervention for object identification and goal understanding, situating them in the middle of the planning autonomy of robots.

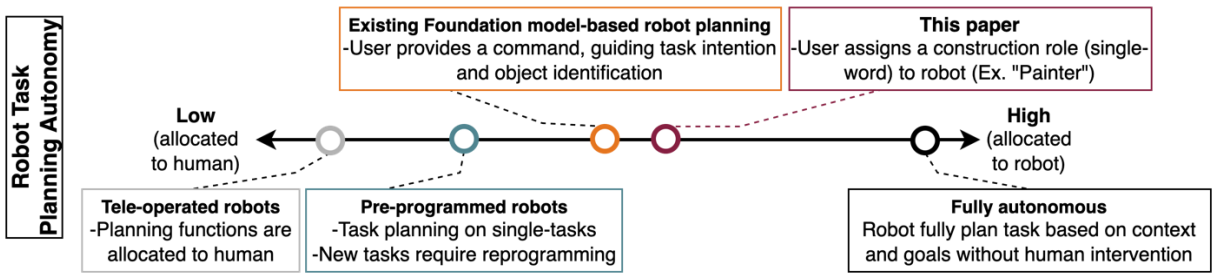

Figure 2 Robot autonomy range in task planning and various paradigms.

However, our study aims to take one step further toward higher autonomy by minimizing the need for detailed human instructions. Rather than providing the robot with sentence commands and instructions, this paper only provides a single word that shows the robot's role in the task. In this method, the robot should autonomously reason and determine the required plan based on the situation without relying on specific keywords in the command. To this end, this research employs a collaborative work between a team of VLMs and LLMs to identify the task goals, best objects for task completion, and effective functions to perform the assigned role. This approach provides a better understanding of multi-AI agent behavior for robot task planning and enables us to move toward somewhere between conditional automation and high automation.

Furthermore, this study can be distinguished from related studies within the context of construction research. One of the few related studies is Kim et al [80], which proposed a framework that uses Building Information Modeling (BIM) and human unstructured data as input to generate robot task planning as output. Despite the valuable findings of this study, our research distinguishes itself in several areas: (1) the study was validated through painting activity in construction and did not demonstrate the model's generalizability for different tasks in construction, which is a significant problem in the field; (2) similar to studies in computer science, this study limited its input to human-specified data with defined situation-specific task instructions, potential safety hazards, and required actions, positioning it in the almost same place; (3) the LLM processing relied on the online ChatGPT API, which posed delays for routine operations and was expensive for all actions. Aligned with recommendations from that study, our research emphasizes local and open-source models, which are more feasible given the high uncertainty in construction settings; (4) the previous study was limited to a single LLM agent, which reported significant issues with reasoning and hallucination [41]. However, one of the objectives of our study is to better understand the capabilities of single and multi-agent systems by comparing them in various construction tasks.

## 3. Method

This section presents four specific design architectures, followed by a discussion of the roles selected for these experiments. Finally, we detail the hardware configurations utilized in this paper.

### 3.1. Framework

Three multi-AI agents along with one single-AI agent are proposed in this section to analyze the performance of the proposed planning systems for autonomous construction robotics. Figure 3 provides an overview of the proposed framework, which is structured over three primary layers: (a) real-world, (b) data management, and (c) task planning designs. Each layer is discussed in detail in the following sections.

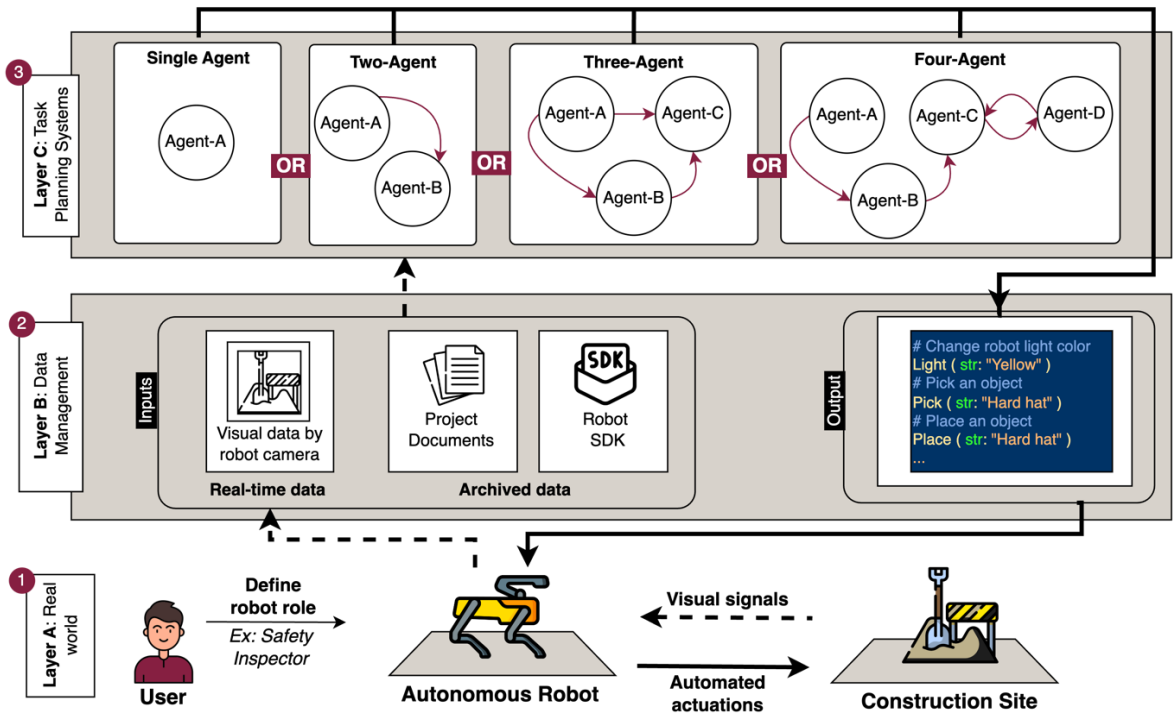

*Figure 3* Overview of the framework.

### 3.2.1. Layer A: Real world entities

The outermost layer, representing the real-world, encompasses the physical entities within the framework: the user, the autonomous robot, and the construction site. At the beginning of the process, the user defines a specific role for the robot to play in the construction site. The rationale behind this approach stems from our discussion in the point of departure section (see section 2.5), where assigning a singular role to the autonomous robot demonstrates a move away from regular command-based inputs. The defined role provided by the user is then transmitted to the robot, which plays a central role in this study. The robot continuously receives real-time signals from the construction site (dashed lines in Figure 3) and subsequently performs actions based on received output from task planning systems.

### 3.2.2. Layer B: Data management

This layer consists of two main data groups: input (dashed lines in Figure 3) and output (solid lines in Figure 3). Input data can be divided into real-time and archived data. Real-time data includes visual data from cameras on robots at the construction site, while archived data consists of project documents such as agreements, safety regulations, schedules, and site-specific information required by the cognition platform. Furthermore, to seamless integration with robot execution systems, the platform requires basic information from robot libraries and documentation, such as Robot Software Development Kits (SDKs) or the Robot Operating System (ROS) Packages, Nodes, and Topics. The output data consists of a task plan generated by our proposed system, which the robot can execute to perform actions based on its assigned role.

### 3.2.3. Layer C: Task planning systems

This layer serves as the core computational module, functioning as the brain of the robot, and plays a central role in achieving the study's objectives. This paper analyzes and compares the output of single-AI agents and three designed multi-AI agents to achieve study objectives. Figure 4 illustrates four different designs proposed and analyzed in this paper. The multi-AI agent systems are designed based on Soar Cognitive Architecture [85], which is a cognitive architecture developed by John Laird in 1983. The goal of this concept is to provide the computational building blocks that are necessary for achieving human-level intelligent systems, including robots. Soar discussed what building blocks are necessary to support human-level generalizable systems. Accordingly, this study sets the Soar theory as its foundation for developing a task planning system with the aim of enabling generalizability and adaptability to construction autonomous robots.

Table 1 characterizes four fundamental computational building blocks: input link, working memory and production system, output link, preference system, and impasse resolution. The Input Link receives and processes sensory data from the environment, creating a representation of the current state. This component is crucial for perceiving the world. The working memory and production system uses the current state along with their knowledge to reason and generate new information and plans. The output link generates muscle or motor commands to interact with the environment, executing plans and taking actions. The preference system and impasse resolution handle conflicts and unexpected situations, improving the system's behavior.

*Table 1* Building blocks of Soar Cognitive Architecture and Potential roles.

| Soar theory roles | Description | Agent Roles |
|---|---|---|
| Input Link | Receive the visual data and describe it | Observer |
| Working Memory and Production System | Generate the robot task planning | Planner |
| Output Link | Generate executable output | Actor |
| Preference System and Impasse Resolution | Proofread and improve the output | Editor |

Based on these building blocks, there are four distinct roles that an adaptable robot task planner is expected to perform. This paper adopted a granular approach to assigning these roles to agents, starting with a compact, single-agent architecture and scaling up to a four-agent system. Figure 4 illustrates these designs; each agent is represented by an avatar to visually distinguish their roles. Maroon arrows indicate the collaboration and communication channels defined between the agents. In addition to inter-agent communication (maroon arrow), each agent operates within its own data collection loop (orange arrow), allowing it to gather and refine specific data before collaboration and communicate it with other agents.

The most compact design is Design A, where a single agent built on a VLM is responsible for generating robot executable task plans. This agent generates executable action plans based on the images it observes. In Design B, there are two agents, the first one acts as an observer and planner, receiving visual data and generating the task plan. The second agent then converts this generated plan into machine-understandable code. Design C further refines the defined roles in Table 1 by introducing three distinct agents, each handling a specific task. The Observer Agent perceives and describes the situation to the Planner Agent, which creates an action plan for the robot. The Actor Agent then receives both the plan and the situation description, converting them into executable commands using the robot's documentation. Finally, in the most granular design, Design D, an additional agent is introduced: the Editor Agent, which proofreads the output from the Actor Agent and provides a refined version of the commands before they are executed.

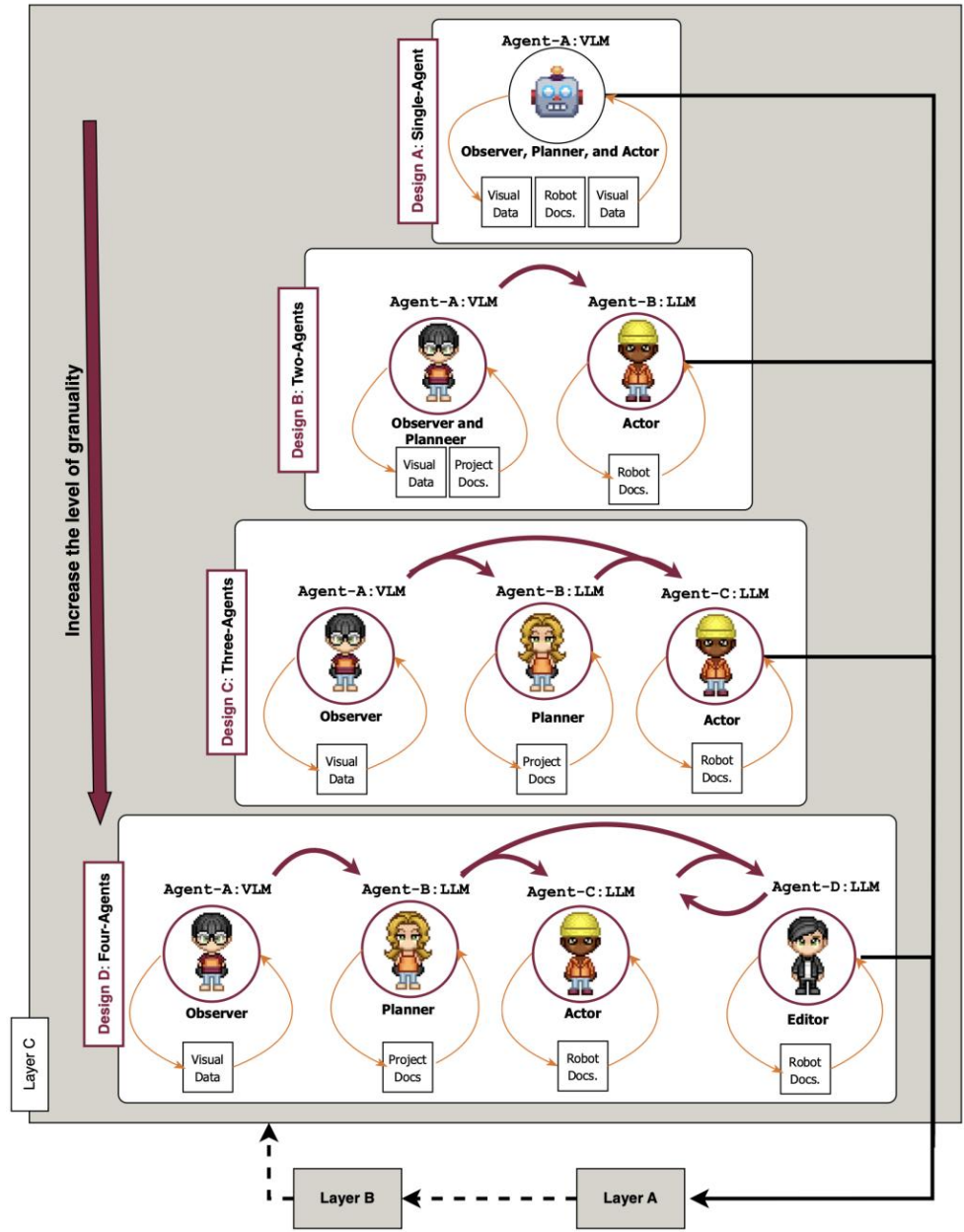

Figure 4 multi-agent system designs and details.

### 4. Experiment Design

To achieve the first study objective, the paper conducted a set of real-world robot experiments to test our models. To ensure reproducibility of this paper, this section discusses the configurations and setting assumed for experiments in the following subsections.

#### 4.1. Activity configurations

This section outlines the rationale behind selecting specific activity for testing our models. To guide our activity selection process, the paper looked into Charles Perrow's "Typology of Technology" [86], which provides a structured approach to categorizing activities based on their inherent characteristics. Figure 4 illustrates the different activity categories of this theory based on two primary dimensions: activity variability and analyzability. Activity variability refers to the frequency of unexpected events or exceptions in the activity, while activity analyzability refers to the extent to which problems can be resolved through analytical processes. These dimensions give rise to four categories of activities: craft, non-routine, routine, and engineering. Next paragraphs discuss the focus of this study for testing the proposed models.

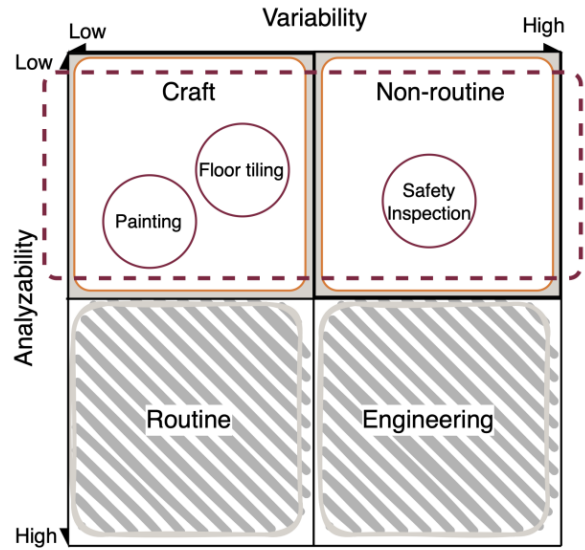

*Figure 5* Activity selection based on Perrow's Typology of Technology.

While tasks with high analyzability (represented in the hashed area of Figure 4) are often the focus of automation in structured environments such as manufacturing, this study is focused on the craft (trades), and non-routine activities (represented in the maroon dashed area), which involve a significant part of daily construction activities [87]. Although prior attempts were focused on creating specific and narrowly defined robots for each task, having a general robot for different tasks remained less automated. Recognizing this gap, our study focuses on testing the capabilities of architectures on three tasks that fall within these domains. In the craft category, the paper selected floor tiling and painting. These tasks are particularly challenging to automate due to their inherent variability; creating rigid if-then statements for such tasks is impractical, as each instance typically requires a unique approach. For the non-routine category, the paper selected the safety inspection task, which involves high variability in site conditions and potential hazards. This task is characterized by low analyzability, as safety hazards are highly context-dependent and difficult to predict or quantify through traditional programming methods [88].

### 4.2. Hardware and software configuration

To conduct experiments, this paper employed a quadruped robot with various attachments. Figure 5 illustrates the robot and various equipment attached to it. This quadruped robot is a Unitree Go2 Edu that is attached to a K1 Robotic Arm and a Robosense Helios32 Lidar. In addition, the experiment was performed on an Ubuntu 20.04 Linux operation laptop with the hardware configuration of Intel Core i9 CPU and Nvidia GeForce RTX 3080 GPU and 32 Gigabyte Memory RAM.

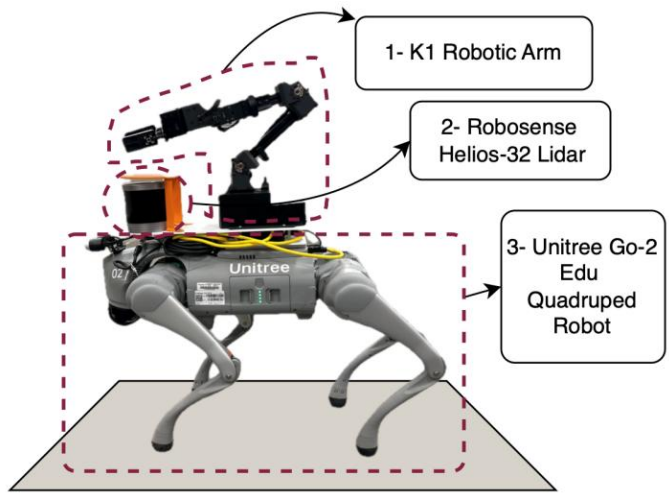

*Figure 6* Robot hardware configurations.

In addition to hardware considerations, the experiment in this study necessitates specific software configurations. At the heart of our automated planning system are different agents, whose core engines are the LLM and VLM models. At the time of writing, GPT-4o represents the state-of-the-art of such models. Despite its capabilities, challenges associated with expenses and using online models (detailed in section 2.5) have led us to decide on local and open-source alternatives. While numerous large-scale LLMs can provide a performance close to GPT-4o, their significant computational and GPU requirements present a substantial barrier. Consequently, our research employs more accessible, lightweight, and cost-effective models [89] [90], better suited to the AEC industry's resource constraints.

During a preliminary one-month testing phase, this study evaluated several lightweight LLMs and VLMs, including LLaVa 1.6-7B, LLaVa 1.6-13B [90], Gemma2 [91], Phi3 [92], Mistral [93], Llama2 7b, and Llama2-13b [94]. Based on these tests, the study selected MiniCPM-2.6-7b [89] as our primary VLM and Llama3-8b [95] as our primary LLM. These selections were informed by an initial pre-testing phase, supervised by two of the authors. While these models proved most suitable for our needs, it is encouraged to assess both existing and emerging models in future studies. Furthermore, this study adopted CrewAI [96] as the primary framework for implementing multi-agent systems. To ensure transparency and support further research, all relevant codes and project details have been made available in a publicly accessible repository[1].

### 4.3. Evaluation

This section outlines the evaluation metrics selected to test our models and meet the objectives of this paper. Well-established metrics are selected from prior related studies [67], [97], including correctness, temporal understanding, and executability. To better address the unique nature of construction tasks, this paper expanded upon these metrics by incorporating additional factors. Temporal understanding, which has often been considered a sub-factor under correctness in previous studies [97], was redefined as a separate metric with its own sub-factors, given its critical role in construction tasks [98], [99]. Additionally, the resource constraints inherent in the construction industry [100], [101], led us to define time and cost as two further metrics, helping to evaluate that the proposed systems are practical and adoptable within real-world construction environments.

---

[1] https://github.com/h-naderi/ai-agent-task-planners

A detailed list of these metrics is provided in Table 2, which includes criteria for assessing model performance across various dimensions. The first metric is Correctness, which measures: (1) how well different architectures use objects within the scene to perform their given roles; (2) how well different architectures understand the (hidden) intention of the scene and perform its responsibility in an exceptional manner; (3) different multi-agent architecture's capability in selecting the most relevant functions for the defined role by the user.

Table 2 Evaluation metrics and its sub-factors

| Metrics | Sub-factors |
| --- | --- |
| Correctness [62], [67], [97] | Correct object usage |
|  | Correct intention prediction |
|  | Function appropriateness |
| Temporal understanding [82], [97] | Correct order of functions and objects |
|  | Semantic understandings |
| Executability [62], [67], [97] | Spatial data hallucinations |
|  | Executable functions based on SDK |
|  | Correct object identification |
| Time | Running time of models |
| Cost | Costs based on token usage |

Next evaluation metric is temporal understanding which assesses the model's capability in maintaining the temporal sequence of actions and its semantic understanding of the tasks. Another key metric is executability, which measures the architecture's capability to resist hallucination, which one of major challenges associated with utilizing LLMs in different applications [102], and to generate executable, actionable plans for robot role operations. This metric measures the model's awareness and handling of spatial data. Moreover, it evaluates how the model adheres to provided SDK functions and avoids generating unsolicited actions beyond documented scope. Additionally, it assesses the model's capability to referencing specific objects seen in the job instead of using general terms, a crucial aspect for minimizing errors during task execution, especially in manipulation phases.

Additionally, the Time metric is introduced to measure the running time of each model, providing insights into the efficiency of the models under different architectures and roles. The Cost metric complements this by estimating expenses based on the token usage of each multi-agent design. This metric helps in evaluating the economic feasibility of deploying these models in real-world scenarios. All of these metrics are benchmarked against GPT-4o, which represents the state of the art at the time of writing this paper.

### 4.4. Painting experiment

This section details the procedure for conducting the painting task experiment. As depicted in Figure 7, the experiment begins with the user (project manager) assigning the role of "Painter Tradesperson" to the robot. Next, the designated workspace was set up for testing the robot in its assigned role. Equipped with a built-in camera, the robot is able to view various painting supplies such as a paint can, primer, painting trays, rollers, a paintbrush, a ladder, and a plywood backdrop.

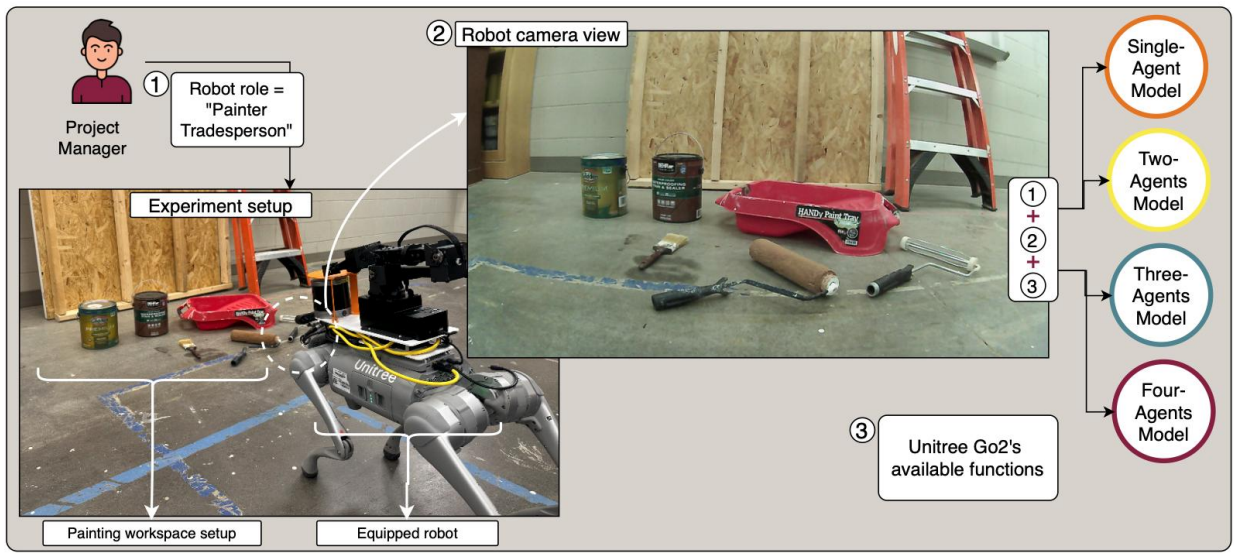

*Figure 7* Painting experiment setup.

The appropriate response to the arrangement is a response that meet all evaluation criteria and subfactors (see Table 2). For example, whether the systems can correctly interpret the scene's hidden intent and execute tasks on plywood rather than the wall. Moreover, it is expected that the appropriate system makes the best use of objects seen in the scene for performing its role. The objects should be identified correctly and with details; instead of "paint can", we expect the appropriate model provides "Behr Painting Can". This kind of response is important for making the model executable for manipulation situations. To better understand the appropriate response from models, a sample output is analyzed completely in the section 5.1.

### 4.5. Safety inspection experiment

Initially, the project manager assigned the role of "Safety Inspection" to the equipped robot in our study. As depicted in Figure 8, the workspace setup demonstrates a worker handling woodwork without wearing the necessary safety gear, having removed his yellow hardhat and safety gloves.

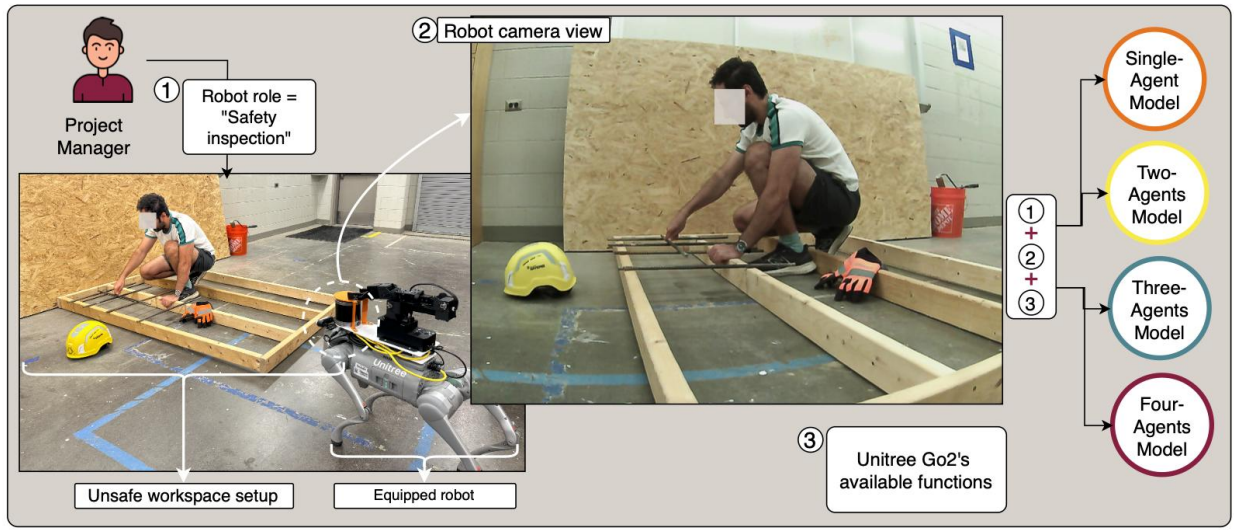

*Figure 8* Safety inspection experiment.

This specific scenario is designed to evaluate whether the robot can identify these safety violations and take appropriate corrective actions. It is anticipated that the models will utilize the appropriate functions in the SDK to pick up the hardhat and safety gloves and inform the worker about the potential danger in the situation. Additionally, the models should demonstrate semantic understanding by mentioning "hard hat" and "safety gloves" first, rather than mentioning the bucket in the scene, as it is less relevant to the robot's role. Furthermore, the models should not mention the woodboard, as it is also not related to the worker's task. Furthermore, the robot captures this scene through its camera, and this visual data, along with other inputs, is integrated into our various multi-agent architectures.

### 4.6. Floor tiling experiment

As illustrated in Figure 9, the floor tiling experiment begins with the project manager assigning the role of "Floor Tiling Tradesperson" to the equipped robot. The workspace for this task has been arranged to assess whether the robot can effectively utilize the provided tiling supplies to complete the tiling work. The job in the experiment is designed to seem ongoing to measure if the proposed models in the robot can understand the appropriate method to complete the task. This setup includes various essential items such as grout, tile spacers, trowels, a rubber mallet, a cleaning sponge, tiles, a surface level, and a tile hammer. It is expected that the models go through this objects in a correct order to complete the task. The robot views this setup through its camera. After capturing the image of the workspace, along with the specific role assignment and robot documentation, this information is transmitted to different architectural models for evaluation.

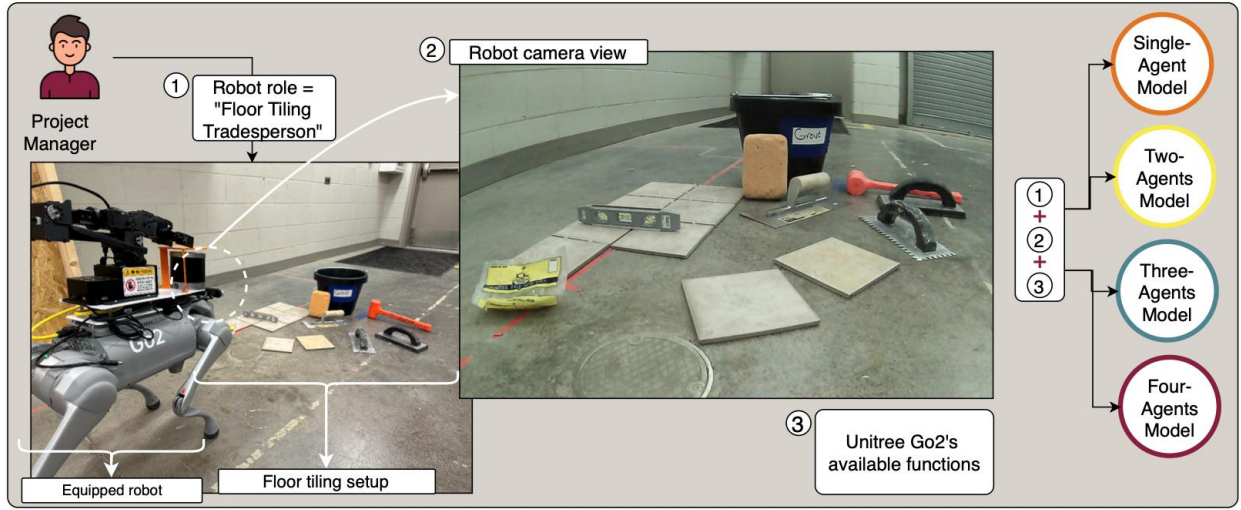

*Figure 9* Floor tiling experiment.

### 4.7. Evaluation method

Unlike conventional machine learning models, which are typically evaluated against predefined correct answers, establishing ground truth responses for LLMs responses, as unstructured text, is not feasible. Consequently, two evaluation methods are widely used: (1) using human evaluators, and (2) using LLMs as evaluators. This paper employed a combination of both approaches to minimize bias in results.

While human evaluation is essential for large-scale prototype deployment, it becomes impractical for experiments with numerous results due to the excessive time and cost involved [103], such as ours with 300 outputs each with 50-100 lines code. To address this, researchers in the LLM domain have recently introduced automatic side-by-side evaluation approaches, such as AutoSxS [104] and LLM-as-a-judge [105]. It is worth noting that LLM-as-a-judge has become a well-established method in LLM domain research as its outputs have been shown to align with human evaluations on recognized benchmarks, affirming the validity of these models [104], [105]. This approach has also been adopted as the primary evaluation method in various studies [106], [107]. This study utilized this approach as a first line of attack in the situation that acquiring human rating data is impractical.

Despite its advantages, several biases and limitations are associated with LLM-as-a-judge, including position bias, self-enhancement bias, and verbosity bias [90]. To mitigate these issues, this paper implemented the following measures: (1) did not use the family of our selected models (Llama 3 and MiniCPM) as judges, but instead used a competitor model (GPT-4o) to minimize self-enhancement bias; (2) Outputs were submitted to the judges anonymously and in random order to reduce position bias; (3) Two of the authors supervised the process, validated the evaluation results, and removed any potential biases from final results.

### 5. Implementation

By implementing proposed single and multi-AI agents in this paper, this section addresses the study objective 1. Furthermore, this section explores some details of implementation and outputs of conducted experiment to lay a foundational understanding of how our proposed system work for addressing next study objectives. It should note that full implementation details, including all codes, prompts, and installed packages are available in the project repository. Aligned with the study scope (refer to section 2.5), our objective is to enable construction robots to be adaptable enough to perform task planning independently for different assigned tasks. To achieve this, it is essential to avoid specifying tasks and actions too narrowly in the prompts given to the models. Accordingly, the prompts used in the four proposed systems don't include specific details, allowing the systems to interpret based on the visual data.

Figure 10 provides some details of the implementation process undertaken specifically in four-agent architecture assigned the role of "Floor Tiling". The first stage of implementation involves defining LLMs and VLMs to be utilized by agents as the core computation core (refer to Figure 10a). In the development of all designs, each agent is assigned a specific role along with a detailed goal and a backstory enhancing its operational understanding and effectiveness (refer to Figure 10b). Responsibilities for each agent are systematically encoded, including description of the responsibility (refer to Figure 10-c1) based on the agent role and an expected output field to guide model outputs. This paper also establishes communication channels through a parameter called "context", which specifies and facilitates interactions between agents (refer to Figure 10-c2).

*Figure 10* Implementation details.

After initializing the agents, each of the three experiments was run 20 times for each specific design, resulting in a total of 240 test results across the four architectural designs and three roles, along with 60 benchmark results from GPT-4o for the same roles. Similar to human teams, agents collaborate with each other through communication channels with the aim of improving the team's output. Figure 11 illustrates some examples of communications that happened within the proposed four-agent architecture tested in the "Floor Tiling Tradesperson" role. This Figure showcases the interactive thought processes of agents and communications, which contribute to formulating the final output. Initially, the Observer agent assesses the floor tiling scene and sends its observations to the Planner agent. The Planner agent is tasked with creating detailed action plans in a language-written format for the robot. Figure 11-a illustrates the executed scenario where the Planner agent identifies a discrepancy based on another agent's observations and adjusts its strategy accordingly. The outputs from these deliberations are then passed to the Actor agent, who converts the plan into executable code. Finally, the Editor agent undertakes a thorough review process, proofreading the Actor's output and cross-verifying it with the Observer's findings and the robot documentation.

This proofread process may include communicating some mistakes to the Actor agent (such as the example in Figure 11-b).

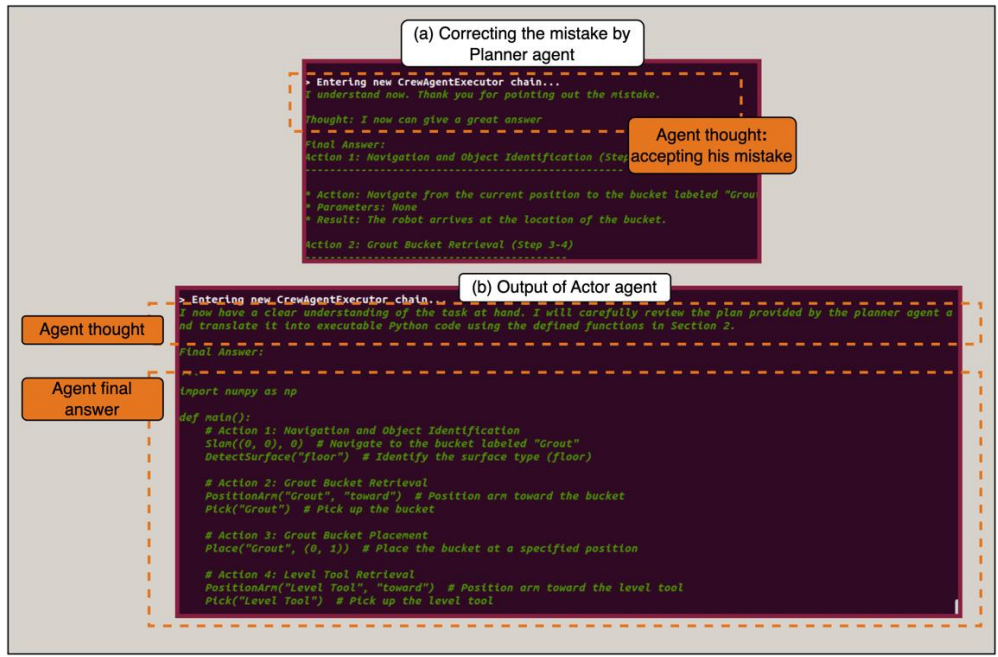

*Figure 11* Agent collaborations and thought process.

### 5.1. Outputs

This section aims to demonstrate some outputs of proposed system to analyze details to ensure the appropriate responses from each designed model. Due to word limit, in this section, four out of 300 outputs are analyzed, and all other outputs can be found under open-source license in the project repository[2]. Figure 12 demonstrates the output of four proposed systems (A to D designs in figure 4), each sequencing task essential for the autonomous robot's assigned role. This figure facilitates comparison of output quality across the different proposed systems. Single-agent architecture output uses the correct functions to navigate, detect surfaces, and apply materials, but it mistakenly applies waterproofing stain to the wall instead of the wooden surface (plywood), which is the intended target (refer to Figure 12-A2). It also picks up the paint tray, which is unnecessary for the current task and wastes time placing it back on the floor (refer to Figure 12-A3 and 4). The paint tray is detected and handled, but it is not used for the painting task. The system should have focused more on objects related to painting the plywood, such as the roller or paintbrush, but they were not utilized. Therefore, the task is partially completed but misdirected towards the wrong surface.

Two-agent architecture output includes identifying surfaces and picking up two paint cans (refer to Figure 12-B2), placing them on the Handy Paint Tray, and applying paint. However, the paint on the floor is incorrect. There's a slight inaccuracy in having the robot detect surfaces and objects multiple times unnecessarily. The paint cans, tray, and roller are all properly used in this output, which shows a little bit better performance compared to single-agent architecture. Three-agent architecture output moves the robot to various target positions, picks up objects like the paintbrush and Behr can, and places them. It applies the Golden primer to the unfinished wooden panel, which is correct (refer to Figure 12-C2). However, it skips important details like ensuring proper coverage or handling the paint roller or tray in a meaningful way. The output uses relevant objects like the paintbrush and Behr can but does not fully engage with other tools that are part of the scene, like the roller and tray.

---

[2] 2 https://github.com/h-naderi/ai-agent-task-planners

In four-agent architecture output, the robot handles the task in a comprehensive way, from picking up the Behr waterproofing stain sealer to applying it to the wooden panel board. It makes good use of the Handy Paint Tray and even changes the light color based on the task. The steps are logical and follow a clear flow from preparing the workspace to applying paint. The Behr stain sealer, paint tray, and paintbrush are all used, which demonstrates a better performance compared to other models. The robot understands the scene well and targets the wooden panel for painting, which is the correct interpretation of the task. However, a critical observation across all models is the tendency to hallucinate spatial data required for certain functions (refer to Figure 12 -A, B, C, D-1). This limitation represents a key area for improvement, which is further elaborated in paragraph four of Section 6.1. It is also important to note that these outputs represent only one set from a total of 60 set implementations. To provide a more comprehensive analysis, the results will be discussed in a broader way in the following section.

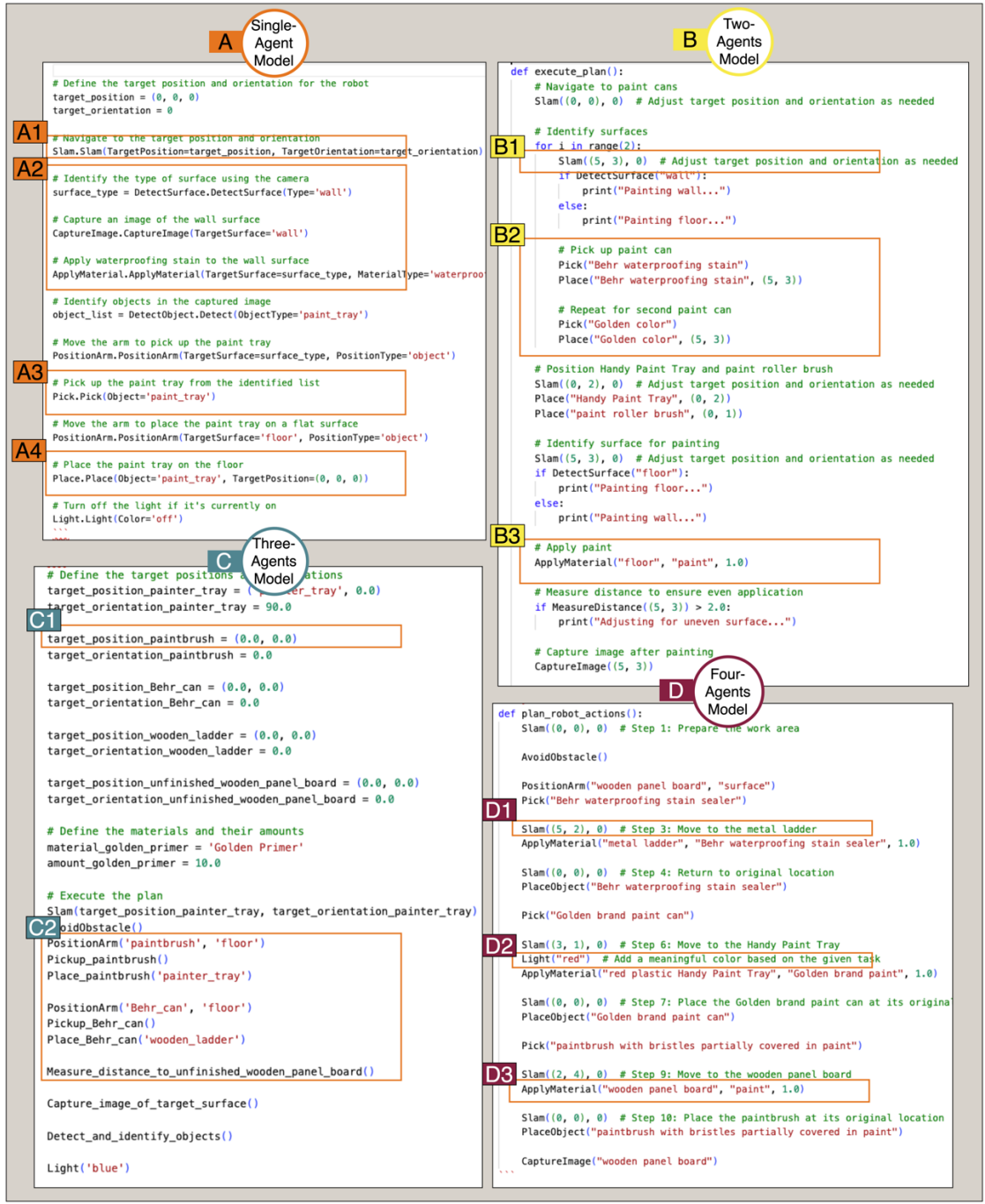

*Figure 12* Sample outputs of four proposed designs tested in the role of "Painter."

## 6. Results

This section compares and analyzes the performance of the proposed task planning systems both relative to each other and in comparison, to GPT-4o. The analysis is initially conducted using three performance metrics: correctness, temporal understanding, and executability, across three predefined roles, namely painter, safety inspector, and floor tiler. The next section extends this analysis by evaluating the same roles based on time and cost metrics. Together, the comparisons and analyses presented here comprehensively address study objective 2.

### 6.1. Performance

Due to the stochastic nature of foundation models, it is crucial to conduct multiple experiments to ensure consistency in the outcome measures. Each of the four designed models represented in Figure 4 was tested 20 times across three roles, resulting in a total of 240 results for each performance metric, including correctness, temporal understanding, and executability. To facilitate comparison among the models and against GPT-4o as the benchmark, all results are presented in 3x3 box plots, as shown in Figure 13. This figure illustrates the performance of four multi-agent architectures across three defined roles for correctness, temporal understanding, and executability metrics. Each row represents the performance score of each role across three metrics, while each column represents the performance score of each of metric across different roles.

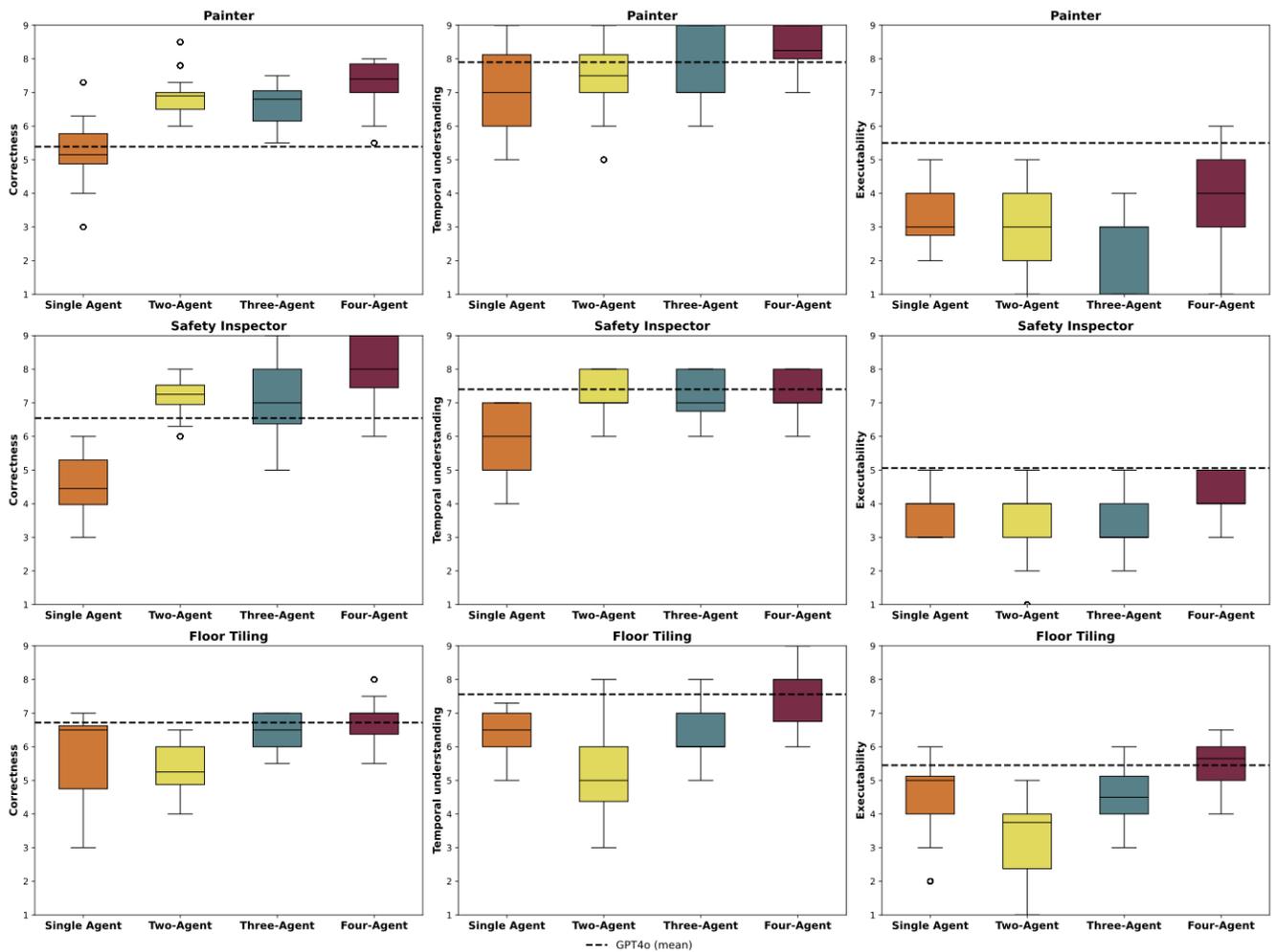

*Figure 13* Study results for three performance metrics across three roles compared to GPT4o results.

The box plots reveal that the Four-Agent architecture consistently performs better than our other designs across all roles and metrics. When compared to GPT4o, the Four-Agent outperforms it in all metrics except for Executability, where the performance of the Four-Agent design is nearly identical to GPT4o (with the exception of the Painter role). In contrast, the Single Agent architecture generally performs the worst, exhibiting greater variability in the interquartile range (IQR) and lower scores across almost all roles and metrics. The Three-Agent architecture demonstrates relatively stable performance, showing moderate medians and variability across most metrics, while the Two-Agent architecture exhibits higher variability and tends to have lower medians in most cases. Overall, the performance of the Three-Agent architecture is comparable to that of GPT4o, while the Four-Agent architecture performs better, and both the Two-Agent and Single Agent architectures perform the worst.

### 6.2. Time

The box plots in Figure 14 illustrate the running time of four designed architectures for task planning across three distinct roles. It's crucial to note that these architectures are being compared against each other, not GPT4o, as they are all installed on the same local machine. Consequently, the comparison with GPT4o in this scenario is irrelevant, as GPT4o is an API-based model that operates on servers. The results presented in the box plots show the running times for different architectures across three roles. In all roles, the trend indicates that the increase in the number of agents is generally associated with an increase in running time, with the Four-Agent architecture demonstrating the highest median times. Notably, the variability of running times differs across roles; in the Painter and Safety Inspector roles, the Single-Agent architecture actually shows higher variability compared to other architectures, as evidenced by the larger IQR. However, for the Floor Tiling role, the variability increases with the number of agents, with the Four-Agent setup having the widest interquartile range. It should be noted that these running-times are dependent on hardware configurations.

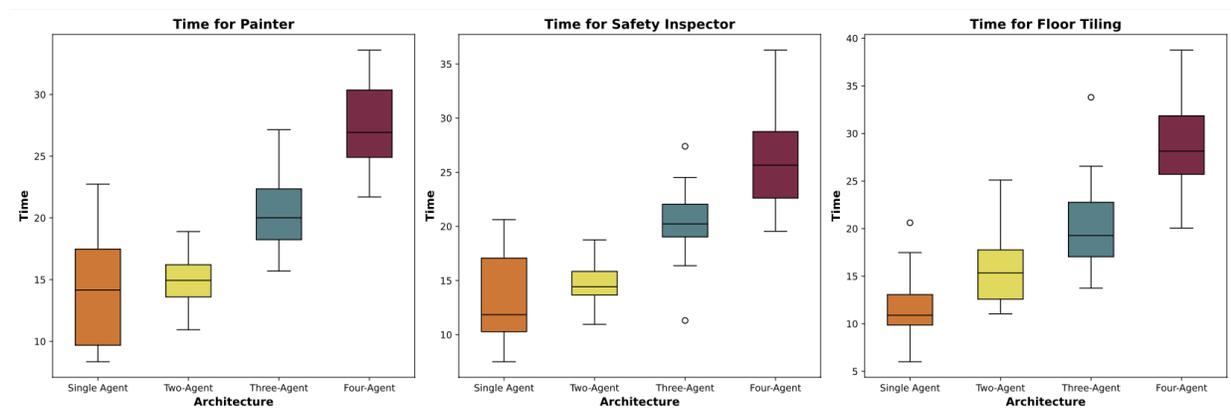

*Figure 14* Test results for running time of different models across three selected roles

### 6.3. Cost

The cost analysis for the multi-agent AI systems was conducted by comparing the per-million-token price for each agent configuration, based on the costs published in the Groq API [108], against the cost of GPT-4o at the time of writing this manuscript. For the single-agent system, which uses a MiniCPM-7B model, the input and output token prices are estimated to be similar to those of similar 7B model, at $0.07 per million tokens. The two-agent configuration utilizes a combination of one MiniCPM-7B model and one Llama 3-8B model, resulting in a cost of $0.12 per million tokens ($0.07 for MiniCPM and $0.05 for the Llama 3-8B model). The three-agent system comprises two Llama 3-8B models and one MiniCPM-7B, amounting to an approximate cost of $0.17 per million tokens. Finally, the four-agent architecture, which includes three Llama 3-8B models and one MiniCPM-7B, incurs a cost of around $0.24 per million tokens. In comparison, GPT-4o, a state-of-the-art model, has a significantly higher cost, estimated at $2.50 per million tokens. This notable difference highlights the cost-efficiency of our proposed architectures, as the most powerful configuration is still approximately ten times cheaper than GPT-4o, while maintaining task-specific performance for zero-shot task planning in robotics (refer to Figure 15).

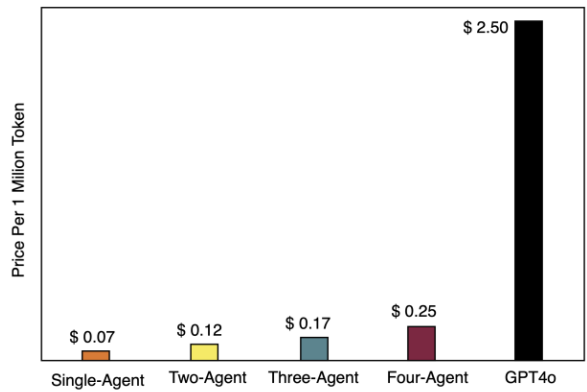

*Figure 15* Study results for cost of models

## 7. Discussion

This section analyzes and interprets the experimental results to derive meaningful insights that address study objective 3. The discussion is organized into three main parts. First, the results are evaluated from an architectural standpoint, focusing on how design choices in agent configurations influence model performance. Second, a detailed analysis of each role is conducted across five key metrics, highlighting the inherent trade-offs between various model configurations. The next subsection explores proposed models' generalizability across three distinct roles. Finally, the section concludes with a discussion of the study's limitations and potential directions for future research.

### 7.1. Agent-centric analysis

Figure 16 illustrates a set of three heatmaps each dedicated to show the average performance of four multi-agent architectures across three defined roles. In this section, this paper focuses our analysis on agent behaviors, where the impact of different defined components on the output performance will be explored.

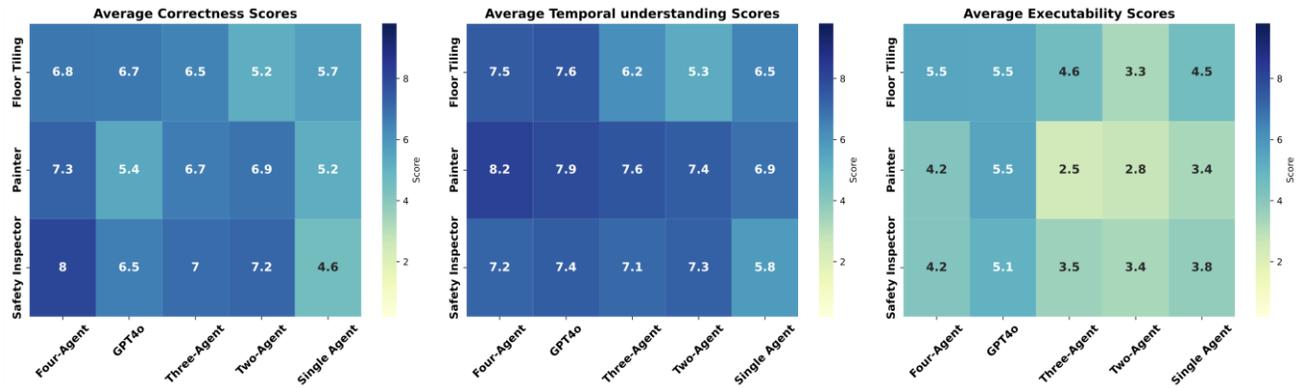

*Figure 16* Average performance of various architectures across three roles.

In the correctness metric (left plot in Figure 16), it is observed that performance generally improves as the number of agents increases (from right to left), where three-agent models reach the level of GPT4o models in most cases. This suggests that the additional role-specific agent and collaboration among them enhances the model's capacity to not only better make use of objects but also better tailor and develop available functions for performing in the given role. Furthermore, this also indicates that models with more collaborative agents have better reasoning capabilities for predicting intention of a situation without exactly mentioning purpose. Another interesting indication is the superior performance of four agents in all roles, most notably in the safety inspector role. This suggests the impact of an additional proofreading agent that helps to ensure task fidelity by revising the translation outputs, thereby reducing errors and enhancing alignment with input instructions.

In terms of temporal understanding (middle plot in Figure 16), a consistent trend of increasing performance is observed as the number of agents increases. The four-agent architecture outperforms the others, achieving the highest score of 8.2 in the "Painter" role. The additional proofreading agent helps the system better align and sequence actions over time, leading to superior temporal coherence. While this trend is significant in both floor tiling and painting roles, increasing the number of agents negatively impacts the temporal understanding of safety inspections. Although an immediate enhancement with the two-agent system can be seen, additional editor and planner agents maintain scores at the GPT4o level performance without further improvement. This suggests that the inherent nature of tasks is an important factor in designing multi-agent architectures, and further studies are needed to better understand the nuances and requirements of different tasks in construction settings.

Regarding executability, a more yellowish range of colors across all architectures and roles can be observed, indicating that this is a major area for improvement. Almost all models are struggling with understanding 3D environments, resulting in hallucinating spatial data in functions. This issue is the primary reason for the lower performance reflected by the yellowish range in this metric (more detailed in the Limitation section). In contrast with the previous two categories, even increasing the number of agents does not help and has had a negative impact on performance. Three-agent architectures reach the same performance level as single-agent architectures, suggesting that the increase in agents may introduce additional complexity and noise, leading to unintended hallucination. However, four-agent systems perform slightly better thanks to the editor agent, which can address some errors and noise. Our proposed models reached 80% of the performance level of the GPT4o model, which is promising when considering the cost of our models. Even GPT4o, as a state-of-the-art commercial model, has reached around 55% of acceptable executability metric, which is still far from an acceptable level. This clearly shows a knowledge gap around the spatial understanding of foundation models, and further studies, such as SPARTUN3D [109], are needed to address this gap.

### 7.2. Trade-off Analysis

This section aims to analyze the relationship and tradeoffs between all performance metrics to give a better understanding of multi-agent systems for robot task planning in construction setting. Understanding these tradeoffs is critical in optimizing both technical and economic aspects of construction automation, ensuring balanced improvements across all criteria. To facilitate meaningful comparisons, it is essential to normalize all metrics to a common scale. The normalization is performed using the following Equation 1, where $x$ is the original value of metric (e.g. cost or time), $x_{min}$ is the minimum value observed, representing the best case, and $x_{max}$ is the maximum value, representing the worst case. This approach ensures that all metrics are normalized on a scale from 0 to 10, where 0 reflects the least desirable outcome and 10 represents the most desirable.

$$Normalized\ Value = 10 \times \left(1 - \frac{x - x_{min}}{x_{max} - x_{min}}\right) \qquad (1)$$

Figure 17 illustrates a set of three different radar plots, each indicating the average score of metrics across three defined roles. Although four-agent architecture generally performs better in terms of three performance metrics, single-agent architecture shows better performance regarding the time and cost metrics. Generally, architectures tend to have a trade-off between time and cost against other three performance metrics, where higher-performing architectures in correctness, temporal understanding, and executability, demonstrate higher cost and time for task completion.

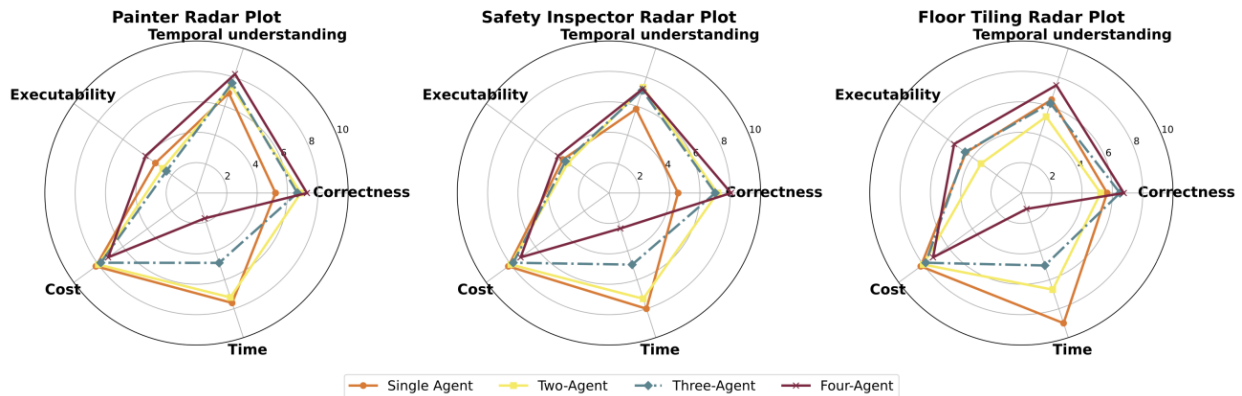

*Figure 17* Overall performance of different architecture across three selected roles.

While four-agent architecture demonstrated higher performance in correctness, temporal understandings, and executability, it needs improvements in cost and time. Regarding the cost, the margin between four-agent architecture and the single-agent architecture is not very significant, both with the cost of $0.25 and $0.07 respectively are much cheaper than GPT4o which costs $2.5 per million. However, when considering the scalability of the proposed systems, four-agent architecture is almost four-times more expensive than single-agent. Authors suggest more comprehensive studies to identify the most efficient architecture for each construction task. Regarding the time, there is big margin between four-agent architecture with average running time of 26.43s compared to single-agent system that complete the task in 11s. Despite the big margin, it should be noted that running times are totally dependent on the hardware configurations, and computing method. To address this challenge along with costs, further studies are suggested to utilize resource-efficient methods [110] to improve the cost and time-efficiency of multi agent systems in construction settings. These studies can be focused on utilizing methods, such as pruning [111], quantization [112], and parameter-based fine-tuning [113], to enhance the current attempt regarding scalability.

### 7.3. Generalizability

This section examines the degree of generalizability achieved by the proposed models. The generalizability can be applied on different variation such as target task, weather condition, and robot morphologies. It is crucial to note that this paper (and this section) aims to understand the generalizability of four proposed architecture across three designed experiments and based on evaluation metrics. The generalizability is often regarded as an inherent characteristic of higher autonomy (refer to Section 2.4) and lacks a widely accepted standard for measurement, making it a qualitative metric. Given this context, a qualitative approach was employed by measuring the arithmetic mean of correctness, temporal understanding, and executability score across three roles and compare the result with the GPT4o results (see Figure 18). If a model demonstrates strong and consistent performance across multiple roles, it can be inferred that a certain degree of generalizability has been attained.

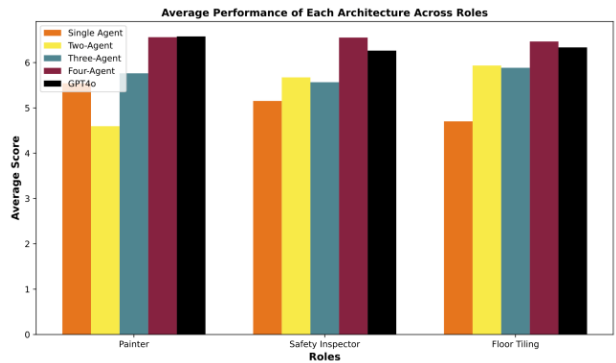

*Figure 18* Average performance of each architecture across three roles

As shown in Figure 18, the three-agent and four-agent architectures demonstrate stable and reliable performance across the three predefined roles. The former showed a consistent score of 5.7 across different roles, while the latter demonstrated a consistent score of 6.5 across all three roles. This consistency underscores the generalizability of these multi-agent systems, highlighting one more step away from traditional task-specific robots, which are often limited to predefined functions. The ability of these architectures to adapt and perform diverse roles with steady efficiency highlights the influence of the multi-agent design capabilities and potential for broader applications. Despite the improved generalizability in three-agent and four-agent architectures, single- and two-agent architectures do not demonstrate a strong generalizability across various roles. This informs a knowledge gap that needs further studies to investigate various design considerations that can improve single- and two-agent generalizability.

## 8. Limitation and Future Studies

Although this paper presents significant findings, like other studies, it is not exempt from limitations. This study presents a novel approach to zero-shot automated robot task planning through lightweight multi-agent systems, but several limitations should be acknowledged. First, our architectures, inspired by the SOAR cognitive architecture, were restricted to four agents with predefined communication channels and roles. While this design proved effective in our experiments, the experiments presented different knowledge gaps that suggest more diverse multi-agent designs with varying communication pathways need to be explored. For example, the impact of specialized agents for coordination, or managing across multiple robots working in parallel should be studies to give a better understanding of AI-team collaborations in construction automation domain.

Additionally, another limitation is that our experiments were limited to three predefined roles based on Perrow's diagram. However, more comprehensive studies of each task group may require some subtle details in different multi-agent designs to its specific needs. Future research should investigate a broader array of construction tasks, focusing on their individual requirements for automation, particularly in multi-agent system design. Understanding the nuances of each task will be crucial in optimizing the model generalizability for different roles and improving the overall robustness of the system in real-world applications.

Moreover, this study utilized a quadruped robot equipped with a pre-installed camera and relied on defined data inputs and SDKs for task execution. While this setup was sufficient for our experiments, future research should explore the deployment of these multi-agent systems in robots with diverse morphologies, such as aerial drones, humanoid robots, or collaborative swarms. For example, future studies can utilize the same multi-agent systems into drones and using drone camera to study their performance in construction task planning in various spatial and environmental conditions, potentially unlocking broader applications in construction automation.

One important limitation of this study lies in the challenges foundation models are struggling with, which is spatial understanding. The current architecture of most foundation models is designed for sequential tokens, making them better suited for cognitive tasks rather than acting tasks that require extensive situated reasoning and direct interaction with 3D data. This restricts their ability to accurately interpret and navigate the typical three-dimensional, dynamic environments of construction sites, potentially impairing the robot's precise movements and adaptability to unforeseen spatial configurations, especially in action level. One key barrier to enhancing the spatial reasoning capabilities of foundation models is the scarcity of 3D data, which remains significantly less available compared to text and image datasets. This issue has been highlighted in other studies, which also emphasize the lack of sufficient 3D data for training foundation models [34]. This highlights the need for future studies to improve the spatial understanding of foundation models.

Another limitation of this study lies in the use of static image data from the robot's camera for task planning. While our approach demonstrated the potential of multi-agent systems to automate various construction tasks, the static nature of the input data does not fully capture the dynamic and ever-changing environment of a real-world construction site. Construction settings are highly fluid, with frequent changes in worker activities, material availability, and environmental conditions. As a result, future systems will need to incorporate real-time data streams, enabling continuous adaptation and real-time task planning. Such systems must not only achieve acceptable levels of automation but also possess the capability to correct their actions rapidly in response to dynamic changes. This adaptive behavior cannot be fully realized using static data inputs, as real-time task adjustments require instant feedback and decision-making to ensure both efficiency and safety in construction tasks. Therefore, future studies should focus on integrating dynamic data inputs, such as live video feeds or sensor data, to enable more responsive and adaptable multi-agent systems.

## 9. Conclusion

This study investigated the use of single- and multi-AI agent systems built over foundation models for enhancing generalizability of robot task planning in autonomous construction robots. The cognitive functions of the robot task planning, were assigned to several agents, collaborating with each other to provide robot action plans. Our experiments tested these multi-agent teams across three critical construction roles: painter, safety inspector, and floor tiler. The results clearly demonstrate that multi-agent architectures, particularly the four-agent system, outperform the state-of-the-art GPT-4o across multiple key metrics. These metrics include task correctness, temporal understanding, and executability, with the four-agent system showing superior performance and generalizability compared to other tested models.

A significant finding is that through the proposed multi-AI agent system, building on lightweight models, we achieved a superior performance compared to the state-of-the-art model, GPT4o. Despite the superior performance, the four-agent architecture is approximately ten times more cost-effective than GPT-4o, making it a potential feasible solution for industries with resource constraints, such as construction. By employing local, open-source LLMs, our system addresses the high costs and dependency on cloud-based services often associated with more powerful closed-source models. This shift towards more affordable and efficient solutions has broad implications for scaling automation in the construction sector and beyond.

Comparing correctness, temporal understanding, and executability metrics across the architectures, the paper finds that adding more agents generally improves correctness and temporal understanding, particularly as tasks become more complex and require more nuanced understanding and sequencing. However, there are diminishing returns, particularly evident in executability scores where the increase of agents led to higher hallucination rates. Nevertheless, it can be generally said that the defined "editor" agent elevates the metrics in the studied cases, highlighting the potential positive effect of an agent with revising and proofreading responsibilities in various situations.

Furthermore, the inclusion of distinct roles, such as the Editor agent, demonstrates the importance of role specialization in improving system performance. This agent significantly contributed to reducing errors and refining task execution, highlighting the value of collaborative, role-specific cognitive processes in complex, real-world scenarios. The role-based agent framework allows for scalability and can be easily extended to accommodate additional tasks or new roles, further enhancing its adaptability to various industries.

The comparative analysis of models in different roles underscored the promising generalizability of the proposed three-agent and four-agent architectures, as demonstrated by their stable and reliable performance across multiple roles. This highlights one further step over traditional task-specific robots, which require considerable reprogramming or retraining for adopting new tasks. The multi-agent design for three- and four-agent systems allowed these systems to handle varied tasks with consistent efficiency, paving the way for broader applications and higher levels of autonomy in the construction environments. However, the relatively weaker performance of the single- and two-agent architectures reveals an important limitation and a knowledge gap. Addressing this gap will require further investigation into design considerations and strategies to enhance generalizability in these simpler configurations

In conclusion, this research lays the groundwork for the broader adoption of multi-agent systems in construction robotics and other environments. The demonstrated cost-effectiveness, modularity, and generalizability make the proposed multi-agent architectures a candidate for future developments in robotic automation. Further research should focus on expanding the range of tasks, exploring additional agent roles, and refining the system for real-time adaptability in highly dynamic environments. These advancements will be crucial for enabling fully autonomous, adaptable robots capable of tackling the diverse challenges of modern construction.

**CRediT authorship**

Hossein Naderi: Conceptualization, Methodology, Software, Validation, Writing – Original Draft.

Alireza Shojaei: Conceptualization, Methodology, Supervision, Writing – Review & Editing.

Lifu Huang, Philip Agee, Kereshmeh Afsari, and Abiola Akanmu: Supervision, Review & Editing.

**References**

[1] P. Teicholz, "Labor-Productivity Declines in the Construction Industry: Causes and Remedies (Another Look) ," AECbytes. Accessed: Jul. 07, 2024. [Online]. Available: https://www.aecbytes.com/viewpoint/2013/issue_67.html

[2] Q. K. Jahanger, D. Trejo, and J. Louis, "Evaluation of field labor and management productivity in the USA construction industry," *Engineering, Construction and Architectural Management*, vol. ahead-of-print, no. ahead-of-print, 2023, doi: 10.1108/ECAM-09-2022-0918/FULL/PDF.

[3] E. Torpey, "Careers in construction: Building opportunity : Career Outlook," U.S. Bureau of Labor Statistics. Accessed: Jul. 07, 2024. [Online]. Available: https://www.bls.gov/careeroutlook/2018/article/careers-in-construction.htm

[4] A. Mohammadi, M. Tavakolan, and Y. Khosravi, "Factors influencing safety performance on construction projects: A review," *Saf Sci*, vol. 109, pp. 382–397, Nov. 2018, doi: 10.1016/J.SSCI.2018.06.017.

[5] BLS, "Construction Labor Productivity ," U.S. Bureau of Labor Statistics. Accessed: Nov. 05, 2024. [Online]. Available: https://www.bls.gov/productivity/highlights/construction-labor-productivity.htm

[6] H. H. Neve, S. Wandahl, S. Lindhard, J. Teizer, and J. Lerche, "Determining the Relationship between Direct Work and Construction Labor Productivity in North America: Four Decades of Insights," *J Constr Eng Manag*, vol. 146, no. 9, p. 04020110, Jun. 2020, doi: 10.1061/(ASCE)CO.1943-7862.0001887.

[7] F. Barbosa and L. Woetzel, "Reinventing construction through a productivity revolution," | McKinsey. Accessed: Nov. 13, 2024. [Online]. Available: https://www.mckinsey.com/capabilities/operations/our-insights/reinventing-construction-through-a-productivity-revolution

[8] L. Larocco and N. Goldberg, "The hard hat job with highest level of open positions ever recorded," CNBC.

[9] BLS, "Number and rate of fatal work injuries, by private industry sector," BLS. Accessed: Nov. 13, 2024. [Online]. Available: https://www.bls.gov/charts/census-of-fatal-occupational-injuries/number-and-rate-of-fatal-work-injuries-by-industry.htm

[10] T. Bock, "The future of construction automation: Technological disruption and the upcoming ubiquity of robotics," *Autom Constr*, vol. 59, pp. 113–121, Nov. 2015, doi: 10.1016/J.AUTCON.2015.07.022.

[11] R. Bogue, "What are the prospects for robots in the construction industry?," *Industrial robot*, vol. 45, no. 1, pp. 1–6, Jan. 2018, doi: 10.1108/IR-11-2017-0194.

[12] T. Bock and T. Linner, *Robot-oriented design*. Cambridge University Press, 2015.

[13] X. Xu and G. B. De Soto, "On-site Autonomous Construction Robots: A review of Research Areas, Technologies, and Suggestions for Advancement," *Proceedings of the 37th International Symposium on Automation and Robotics in Construction, ISARC 2020: From Demonstration to Practical Use - To New Stage of Construction Robot*, pp. 385–392, 2020, doi: 10.22260/ISARC2020/0055.

[14] L. Zeng, S. Guo, J. Wu, and B. Markert, "Autonomous mobile construction robots in built environment: A comprehensive review," *Developments in the Built Environment*, vol. 19, p. 100484, Oct. 2024, doi: 10.1016/J.DIBE.2024.100484.

[15] N. Melenbrink, J. Werfel, and A. Menges, "On-site autonomous construction robots: Towards unsupervised building," *Autom Constr*, vol. 119, p. 103312, Nov. 2020, doi: 10.1016/J.AUTCON.2020.103312.

[16] A. Ojha, M. Habibnezhad, H. Jebelli, and R. Leicht, "Barrier Analysis of Effective Implementation of Robotics in the Construction Industry," *Construction Research Congress 2022: Computer Applications, Automation, and Data Analytics - Selected Papers from Construction Research Congress 2022*, vol. 2-B, pp. 661–669, 2022, doi: 10.1061/9780784483961.069.

[17] P. Pradhananga, M. ElZomor, and G. Santi Kasabdji, "Identifying the Challenges to Adopting Robotics in the US Construction Industry," *J Constr Eng Manag*, vol. 147, no. 5, p. 05021003, May 2021, doi: 10.1061/(ASCE)CO.1943-7862.0002007/ASSET/085304FC-53CF-4444-9BBD-7E90360F1618/ASSETS/IMAGES/LARGE/FIGURE5.JPG.

[18] M. Y. Bin Yahya, Y. L. Hui, A. B. Md. Yassin, R. Omar, R. O. anak Robin, and N. Kasim, "The Challenges of the Implementation of Construction Robotics Technologies in the Construction," *MATEC Web of Conferences*, vol. 266, p. 05012, 2019, doi: 10.1051/MATECCONF/201926605012.

[19] X. Wang, S. Wang, C. C. Menassa, V. R. Kamat, and W. McGee, "Automatic high-level motion sequencing methods for enabling multi-tasking construction robots," *Autom Constr*, vol. 155, p. 105071, Nov. 2023, doi: 10.1016/J.AUTCON.2023.105071.

[20] J. Wang, W. Liang, J. Yang, S. Wang, and Z.-X. Yang, "An adaptive image enhancement approach for safety monitoring robot under insufficient illumination condition," *Comput Ind*, vol. 147, p. 103862, May 2023, doi: 10.1016/J.COMPIND.2023.103862.

[21] Hilti, "Jaibot Drilling Robot ," 2023. Accessed: Nov. 13, 2024. [Online]. Available: https://www.hilti.com/content/hilti/W1/US/en/business/business/trends/jaibot.html

[22] Construction robotic, "SAM - Bricklaying made simpler and safer. ," 2023. Accessed: Nov. 13, 2024. [Online]. Available: https://www.construction-robotics.com/sam-2/

[23] X. Wang, H. Yu, W. McGee, C. C. Menassa, and V. R. Kamat, "Enabling Building Information Model-driven human-robot collaborative construction workflows with closed-loop digital twins," *Comput Ind*, vol. 161, p. 104112, Oct. 2024, doi: 10.1016/J.COMPIND.2024.104112.

[24] S. Metvaei, K. Aghajamali, Q. Chen, and Z. Lei, "Developing a BIM-enabled robotic manufacturing framework to facilitate mass customization of prefabricated buildings," *Comput Ind*, vol. 164, p. 104201, Jan. 2025, doi: 10.1016/J.COMPIND.2024.104201.

[25] A. I. Karoly, P. Galambos, J. Kuti, and I. J. Rudas, "Deep Learning in Robotics: Survey on Model Structures and Training Strategies," *IEEE Trans Syst Man Cybern Syst*, vol. 51, no. 1, pp. 266–279, Jan. 2021, doi: 10.1109/TSMC.2020.3018325.

[26] J. Manuel Davila Delgado and L. Oyedele, "Robotics in construction: A critical review of the reinforcement learning and imitation learning paradigms," *Advanced Engineering Informatics*, vol. 54, p. 101787, Oct. 2022, doi: 10.1016/J.AEI.2022.101787.

[27] J. Manuel Davila Delgado and L. Oyedele, "Robotics in construction: A critical review of the reinforcement learning and imitation learning paradigms," *Advanced Engineering Informatics*, vol. 54, p. 101787, Oct. 2022, doi: 10.1016/J.AEI.2022.101787.

[28] F. Crespin-Mazet and P. Portier, "The reluctance of construction purchasers towards project partnering," *Journal of Purchasing and Supply Management*, vol. 16, no. 4, pp. 230–238, Dec. 2010, doi: 10.1016/J.PURSUP.2010.06.001.

[29] E. Poirier, D. Forgues, and S. Staub-French, "Collaboration through innovation: implications for expertise in the AEC sector," *Construction Management and Economics*, vol. 34, no. 11, pp. 769–789, Nov. 2016, doi: 10.1080/01446193.2016.1206660.

[30] M. Seo, S. Gupta, and Y. Ham, "Evaluation of Work Performance, Task Load, and Behavior Changes on Time-Delayed Teleoperation Tasks in Space Construction," *Construction Research Congress 2024, CRC 2024*, vol. 1, pp. 89–98, 2024, doi: 10.1061/9780784485262.010.

[31] M. Ghallab, D. Nau, and P. Traverso, "Automated Planning and Acting," *Automated Planning and Acting*, pp. 1–354, Aug. 2016, doi: 10.1017/CBO9781139583923.

[32] M. Leonetti, L. Iocchi, and P. Stone, "A synthesis of automated planning and reinforcement learning for efficient, robust decision-making," *Artif Intell*, vol. 241, pp. 103–130, Dec. 2016, doi: 10.1016/J.ARTINT.2016.07.004.

[33] Y. Hayamizu, S. Amiri, K. Chandan, K. Takadama, and S. Zhang, "Guiding Robot Exploration in Reinforcement Learning via Automated Planning," *International Conference on Automated Planning and Scheduling*, vol. 2021-August, pp. 625–633, 2021, doi: 10.1609/ICAPS.V31I1.16011.

[34] R. Firoozi et al., "Foundation Models in Robotics: Applications, Challenges, and the Future," 2023.

[35] X. Xiao et al., "Robot Learning in the Era of Foundation Models: A Survey," Nov. 2023, Accessed: Feb. 25, 2024. [Online]. Available: https://arxiv.org/abs/2311.14379v1

[36] R. Bommasani et al., "On the Opportunities and Risks of Foundation Models," Aug. 2021, Accessed: Mar. 27, 2024. [Online]. Available: https://arxiv.org/abs/2108.07258v3

[37] A. Vaswani et al., "Attention Is All You Need," *Adv Neural Inf Process Syst*, vol. 2017-December, pp. 5999–6009, Jun. 2017, Accessed: Feb. 16, 2024. [Online]. Available: https://arxiv.org/abs/1706.03762v7

[38] IBM, "Foundation Models," IBM Research. Accessed: Mar. 27, 2024. [Online]. Available: https://research.ibm.com/topics/foundation-models

[39] J. Achiam et al., "GPT-4 Technical Report," *ArXiv*, Mar. 2023, Accessed: Feb. 16, 2024. [Online]. Available: https://arxiv.org/abs/2303.08774v4

[40] H. Touvron et al., "LLaMA: Open and Efficient Foundation Language Models," Feb. 2023, Accessed: Mar. 28, 2024. [Online]. Available: https://arxiv.org/abs/2302.13971v1

[41] F. Xu, Q. Lin, J. Han, T. Zhao, J. Liu, and E. Cambria, "Are Large Language Models Really Good Logical Reasoners? A Comprehensive Evaluation and Beyond," Jun. 2023, Accessed: Jul. 08, 2024. [Online]. Available: https://arxiv.org/abs/2306.09841v3

[42] Y. Du, S. Li, A. Torralba, J. B. Tenenbaum, I. Mordatch, and G. Brain, "Improving Factuality and Reasoning in Language Models through Multiagent Debate," May 2023, Accessed: Jul. 08, 2024. [Online]. Available: https://arxiv.org/abs/2305.14325v1

[43] G. Li et al., "CAMEL: Communicative Agents for 'Mind' Exploration of Large Language Model Society," Mar. 2023, Accessed: Jul. 08, 2024. [Online]. Available: https://arxiv.org/abs/2303.17760v2

[44] T. Liang et al., "Encouraging Divergent Thinking in Large Language Models through Multi-Agent Debate," May 2023, Accessed: Jul. 08, 2024. [Online]. Available: https://arxiv.org/abs/2305.19118v2

[45] Y. qian Jiang, S. qi Zhang, P. Khandelwal, and P. Stone, "Task planning in robotics: an empirical comparison of PDDL- and ASP-based systems," *Frontiers of Information Technology and Electronic Engineering*, vol. 20, no. 3, pp. 363–373, Mar. 2019, doi: 10.1631/FITEE.1800514/METRICS.

[46] E. Karpas and D. Magazzeni, "Automated Planning for Robotics," *Annu Rev Control Robot Auton Syst*, vol. 3, no. Volume 3, 2020, pp. 417–439, May 2020, doi: 10.1146/ANNUREV-CONTROL-082619-100135/CITE/REFWORKS.

[47] R. E. Fikes and N. J. Nilsson, "Strips: A new approach to the application of theorem proving to problem solving," *Artif Intell*, vol. 2, no. 3–4, pp. 189–208, Dec. 1971, doi: 10.1016/0004-3702(71)90010-5.

[48] D. McDermott *et al.*, "PDDL-the planning domain definition language," 1998.

[49] Y. qian Jiang, S. qi Zhang, P. Khandelwal, and P. Stone, "Task planning in robotics: an empirical comparison of PDDL- and ASP-based systems," *Frontiers of Information Technology & Electronic Engineering*, vol. 20, no. 3, pp. 363–373, Mar. 2018, doi: 10.1631/FITEE.1800514.

[50] A. Coles and A. Coles, "PDDL+ Planning with Events and Linear Processes," *International Conference on Automated Planning and Scheduling*, vol. 2014-January, no. January, pp. 74–82, 2014, doi: 10.1609/ICAPS.V24I1.13647.

[51] N. Kushmerick, S. Hanks, and D. S. Weld, "An algorithm for probabilistic planning," *Artif Intell*, vol. 76, no. 1–2, pp. 239–286, Jul. 1995, doi: 10.1016/0004-3702(94)00087-H.

[52] C. C. White and D. J. White, "Markov decision processes," *Eur J Oper Res*, vol. 39, no. 1, pp. 1–16, Mar. 1989, doi: 10.1016/0377-2217(89)90348-2.

[53] M. T. J. Spaan, "Partially Observable Markov Decision Processes," *Adaptation, Learning, and Optimization*, vol. 12, pp. 387–414, 2012, doi: 10.1007/978-3-642-27645-3_12.

[54] G. Konidaris, L. P. Kaelbling, and T. Lozano-Perez, "From Skills to Symbols: Learning Symbolic Representations for Abstract High-Level Planning," *Journal of Artificial Intelligence Research*, vol. 61, pp. 215–289, Jan. 2018, doi: 10.1613/JAIR.5575.

[55] C. Diehl, J. Adamek, M. Krüger, F. Hoffmann, and T. Bertram, "Differentiable Constrained Imitation Learning for Robot Motion Planning and Control," *arXiv.org*, 2022, doi: 10.48550/ARXIV.2210.11796.

[56] T. Kojima, S. S. Gu, M. Reid, Y. Matsuo, and Y. Iwasawa, "Large Language Models are Zero-Shot Reasoners," *Adv Neural Inf Process Syst*, vol. 35, May 2022, Accessed: Dec. 15, 2024. [Online]. Available: https://arxiv.org/abs/2205.11916v4

[57] Y. Jin *et al.*, "RobotGPT: Robot Manipulation Learning from ChatGPT," *IEEE Robot Autom Lett*, vol. 9, no. 3, pp. 2543–2550, Dec. 2023, doi: 10.1109/LRA.2024.3357432.

[58] Y. Cui, S. Niekum, A. Gupta, V. Kumar, and A. Rajeswaran, "Can Foundation Models Perform Zero-Shot Task Specification For Robot Manipulation?," *Proc Mach Learn Res*, vol. 168, pp. 893–905, 2022.

[59] S. Ouyang and L. Li, "AutoPlan: Automatic Planning of Interactive Decision-Making Tasks With Large Language Models", Accessed: Jun. 08, 2024. [Online]. Available: https://github.com/owaski/AutoPlan.

[60] B. Xie *et al.*, "ChatGPT for Robotics: A New Approach to Human-Robot Interaction and Task Planning," *Lecture Notes in Computer Science (including subseries Lecture Notes in Artificial Intelligence and Lecture Notes in Bioinformatics)*, vol. 14271 LNAI, pp. 365–376, 2023, doi: 10.1007/978-981-99-6495-6_31.

[61] Y. Ye, H. You, and J. Du, "Improved Trust in Human-Robot Collaboration With ChatGPT," *IEEE Access*, vol. 11, pp. 55748–55754, 2023, doi: 10.1109/ACCESS.2023.3282111.

[62] B. Zhang and H. Soh, "Large Language Models as Zero-Shot Human Models for Human-Robot Interaction," *IEEE International Conference on Intelligent Robots and Systems*, pp. 7961–7968, Jan. 2023, doi: 10.1109/IROS55552.2023.10341488.

[63] H. Naderi, A. Shojaei, and L. Huang, "Foundation Models for Autonomous Robots in Unstructured Environments," Jul. 2024, Accessed: Aug. 06, 2024. [Online]. Available: https://arxiv.org/abs/2407.14296v2

[64] Q. Wu *et al.*, "AutoGen: Enabling Next-Gen LLM Applications via Multi-Agent Conversation," Aug. 2023, Accessed: Aug. 06, 2024. [Online]. Available: https://arxiv.org/abs/2308.08155v2

[65] G. Li *et al.*, "CAMEL: Communicative Agents for 'Mind' Exploration of Large Language Model Society," Mar. 2023, Accessed: Aug. 06, 2024. [Online]. Available: https://arxiv.org/abs/2303.17760v2

[66] Y. Du, S. Li, A. Torralba, J. B. Tenenbaum, I. Mordatch, and G. Brain, "Improving Factuality and Reasoning in Language Models through Multiagent Debate," May 2023, Accessed: Aug. 06, 2024. [Online]. Available: https://arxiv.org/abs/2305.14325v1

[67] S. S. Kannan, V. L. N. Venkatesh, and B.-C. Min, "SMART-LLM: Smart Multi-Agent Robot Task Planning using Large Language Models," Sep. 2023, Accessed: Aug. 06, 2024. [Online]. Available: https://arxiv.org/abs/2309.10062v2

[68] T. Liang *et al.*, "Encouraging Divergent Thinking in Large Language Models through Multi-Agent Debate," May 2023, Accessed: Aug. 06, 2024. [Online]. Available: https://arxiv.org/abs/2305.19118v3

[69] H. Ardiny, S. Witwicki, and F. Mondada, "Construction automation with autonomous mobile robots: A review," *International Conference on Robotics and Mechatronics, ICROM 2015*, pp. 418–424, Dec. 2015, doi: 10.1109/ICROM.2015.7367821.

[70] Caterpillar, "Automation & Autonomy: What's the Difference? ," Caterpillar. Accessed: Oct. 31, 2024. [Online]. Available: https://www.cat.com/en_US/articles/ci-articles/automation-autonomy-whats-the-difference.html

[71] O. Moselhi, P. Fazio, and S. Hason, "Automation of concrete slab-on-grade construction," *Journal of Construction Engineering and Management-asce*, vol. 118, no. 4, pp. 731–748, Dec. 1992, doi: 10.1061/(ASCE)0733-9364(1992)118:4(731).

[72] V. S. S. Kumar, I. Prasanthi, and A. Leena, "Robotics and automation in construction industry," *Proceedings of the AEI 2008 Conference - AEI 2008: Building Integration Solutions*, vol. 328, 2008, doi: 10.1061/41002(328)3.

[73] E. J. de Visser, R. Pak, and T. H. Shaw, "From 'automation' to 'autonomy': the importance of trust repair in human–machine interaction," *Ergonomics*, vol. 61, no. 10, pp. 1409–1427, Oct. 2018, doi: 10.1080/00140139.2018.1457725.

[74] W. Xu, "From automation to autonomy and autonomous vehicles: Challenges and Opportunities for Human-Computer Interaction," *Interactions*, vol. 28, no. 1, pp. 48–53, Dec. 2020, doi: 10.1145/3434580/ASSET/A1369F45-92F6-468E-87DC-814F1EB0E0CD/ASSETS/3434580.FP.PNG.

[75] SAE, "Taxonomy and Definitions for Terms Related to Driving Automation Systems for On-Road Motor Vehicles ," Society of Automative Engineers. Accessed: Apr. 30, 2024. [Online]. Available: https://www.sae.org/standards/content/j3016_202104/

[76] J. Beer, A. Fisk, and W. Rogers, "Toward a framework for levels of robot autonomy in human-robot interaction," *J Hum Robot Interact*, vol. 3, no. 2, p. 74, Jul. 2014, doi: 10.5898/JHRI.3.2.BEER.

[77] R. Parasuraman, T. B. Sheridan, and C. D. Wickens, "A model for types and levels of human interaction with automation," *IEEE Transactions on Systems, Man, and Cybernetics Part A:Systems and Humans.*, vol. 30, no. 3, pp. 286–297, 2000, doi: 10.1109/3468.844354.

[78] N. Correll, "Introduction to Autonomous Control," *University of Colorado Boulder*, pp. 1–226, 2016, Accessed: Nov. 04, 2024. [Online]. Available: https://open.umn.edu/opentextbooks/textbooks/316

[79] S. Thrun, "Toward a Framework for Human-Robot Interaction," *Hum Comput Interact*, 2004, doi: 10.1080/07370024.2004.9667338.

[80] R. R. Murphy, *Introduction to AI robotics, Second Edition*. 2019.

[81] I. Singh et al., "ProgPrompt: Generating Situated Robot Task Plans using Large Language Models," Sep. 2022, Accessed: Jun. 17, 2024. [Online]. Available: https://arxiv.org/abs/2209.11302v1

[82] M. Ahn et al., "Do As I Can, Not As I Say: Grounding Language in Robotic Affordances," *Proc Mach Learn Res*, vol. 205, pp. 287–318, Apr. 2022, Accessed: Jun. 17, 2024. [Online]. Available: https://arxiv.org/abs/2204.01691v2

[83] W. Huang, P. Abbeel, D. Pathak, and I. Mordatch, "Language Models as Zero-Shot Planners: Extracting Actionable Knowledge for Embodied Agents," *Proc Mach Learn Res*, vol. 162, pp. 9118–9147, Jan. 2022, Accessed: Jun. 16, 2024. [Online]. Available: https://arxiv.org/abs/2201.07207v2

[84] K. Kim, M. Ivashchenko, P. Ghimire, and P.-C. Huang, "Context-Aware and Adaptive Task Planning for Autonomous Construction Robots Through Llm-Robot Communication", doi: 10.2139/SSRN.4827728.

[85] J. E. . LAIRD, "SOAR COGNITIVE ARCHITECTURE," 2019.

[86] C. Perrow, "A FRAMEWORK FOR THE COMPARATIVE ANALYSIS OF ORGANIZATIONS," *Am Sociol Rev*, vol. 32, no. 2, p. 194, Apr. 1967, doi: 10.2307/2091811.

[87] P. Agee, "A Macroergonomics Path to Human-centered, Adaptive Buildings," Sep. 26, 2019, *Virginia Tech*. Accessed: Mar. 27, 2024. [Online]. Available: http://hdl.handle.net/10919/102751

[88] J. H. Ryu, T. McFarland, B. Banting, C. T. Haas, and E. Abdel-Rahman, "Health and productivity impact of semi-automated work systems in construction," *Autom Constr*, vol. 120, Dec. 2020, doi: 10.1016/J.AUTCON.2020.103396.

[89] Y. Yao et al., "MiniCPM-V: A GPT-4V Level MLLM on Your Phone," Aug. 2024, Accessed: Sep. 23, 2024. [Online]. Available: https://arxiv.org/abs/2408.01800v1

[90] H. Liu, C. Li, Q. Wu, and Y. J. Lee, "Visual Instruction Tuning," *Adv Neural Inf Process Syst*, vol. 36, Apr. 2023, Accessed: Sep. 23, 2024. [Online]. Available: https://arxiv.org/abs/2304.08485v2

[91] G. Team et al., "Gemma 2: Improving Open Language Models at a Practical Size," Jul. 2024, Accessed: Sep. 23, 2024. [Online]. Available: https://arxiv.org/abs/2408.00118v2

[92] M. Abdin et al., "Phi-3 Technical Report: A Highly Capable Language Model Locally on Your Phone," Apr. 2024, Accessed: Sep. 23, 2024. [Online]. Available: https://arxiv.org/abs/2404.14219v4

[93] A. Q. Jiang *et al.*, "Mistral 7B," Oct. 2023, Accessed: Sep. 23, 2024. [Online]. Available: https://arxiv.org/abs/2310.06825v1
[94] H. Touvron *et al.*, "Llama 2: Open Foundation and Fine-Tuned Chat Models," Jul. 2023, Accessed: Sep. 23, 2024. [Online]. Available: https://arxiv.org/abs/2307.09288v2
[95] A. Dubey *et al.*, "The Llama 3 Herd of Models," Jul. 2024, Accessed: Sep. 23, 2024. [Online]. Available: https://arxiv.org/abs/2407.21783v2
[96] CrewAI, "Documentations," CrewAI. Accessed: Oct. 15, 2024. [Online]. Available: https://docs.crewai.com/introduction
[97] R. Wang, Z. Yang, Z. Zhao, X. Tong, Z. Hong, and K. Qian, "LLM-based Robot Task Planning with Exceptional Handling for General Purpose Service Robots," May 2024, Accessed: Oct. 06, 2024. [Online]. Available: https://arxiv.org/abs/2405.15646v1
[98] V. Grosjean and P. Terrier, "Temporal awareness: pivotal in performance?," *Ergonomics*, vol. 42, no. 11, pp. 1443–1456, Nov. 1999, doi: 10.1080/001401399184802.
[99] G. Lucko and A. Mukherjee, "Temporal perspectives in construction simulation modeling," *Online World Conference on Soft Computing in Industrial Applications*, pp. 3257–3268, 2013, doi: 10.1109/WSC.2013.6721691.
[100] G. Heravi and S. Moridi, "Resource-Constrained Time-Cost Tradeoff for Repetitive Construction Projects," *KSCE Journal of Civil Engineering*, vol. 23, no. 8, pp. 3265–3274, Aug. 2019, doi: 10.1007/S12205-019-0151-X.
[101] M. J. Taheri Amiri, F. R. Haghighi, E. Eshtehardian, and O. Abessi, "Multi-project Time-cost Optimization in Critical Chain with Resource Constraints," *KSCE Journal of Civil Engineering*, vol. 22, no. 10, pp. 3738–3752, Oct. 2018, doi: 10.1007/S12205-017-0691-X.
[102] Z. Ji, T. Yu, Y. Xu, N. Lee, E. Ishii, and P. Fung, "Towards Mitigating LLM Hallucination via Self Reflection," pp. 1827–1843, Dec. 2023, doi: 10.18653/V1/2023.FINDINGS-EMNLP.123.
[103] M. Kahng *et al.*, "LLM Comparator: Visual Analytics for Side-by-Side Evaluation of Large Language Models," *Proceedings of ACM Conference (Conference'17)*, vol. 1, Feb. 2024, Accessed: Feb. 19, 2024. [Online]. Available: https://arxiv.org/abs/2402.10524v1
[104] Google Cloud, "AutoSxS: Perform automatic side-by-side evaluation." Accessed: Feb. 19, 2024. [Online]. Available: https://cloud.google.com/vertex-ai/docs/generative-ai/models/side-by-side-eval#aggregate-metrics
[105] L. Zheng *et al.*, "Judging LLM-as-a-Judge with MT-Bench and Chatbot Arena," Jun. 2023, Accessed: Feb. 19, 2024. [Online]. Available: https://arxiv.org/abs/2306.05685v4
[106] E. Guthrie, D. Levy, and G. del Carmen, "The Operating and Anesthetic Reference Assistant (OARA): A fine-tuned large language model for resident teaching," *The American Journal of Surgery*, Feb. 2024, doi: 10.1016/J.AMJSURG.2024.02.016.
[107] Z. Li, B. Peng, P. He, M. Galley, J. Gao, and X. Yan, "Guiding Large Language Models via Directional Stimulus Prompting," in *Advances in Neural Information Processing Systems*, NeurIPS Proceedings, Feb. 2024. Accessed: Feb. 20, 2024. [Online]. Available: https://github.com/Leezekun/Directional-Stimulus-Prompting
[108] Groq, "On-demand Pricing for Tokens-as-a-Service." Accessed: Oct. 09, 2024. [Online]. Available: https://groq.com/pricing/
[109] Y. Zhang, Z. Xu, Y. Shen, P. Kordjamshidi, and L. Huang, "SPARTUN3D: Situated Spatial Understanding of 3D World in Large Language Models," Oct. 2024, Accessed: Oct. 15, 2024. [Online]. Available: https://arxiv.org/abs/2410.03878v1

[110] Z. Wan *et al.*, "Efficient Large Language Models: A Survey," Dec. 2023, Accessed: Oct. 13, 2024. [Online]. Available: https://arxiv.org/abs/2312.03863v4
[111] Z. Ankner, C. Blakeney, K. Sreenivasan, M. Marion, M. L. Leavitt, and M. Paul, "Perplexed by Perplexity: Perplexity-Based Data Pruning With Small Reference Models," May 2024, Accessed: Oct. 13, 2024. [Online]. Available: https://arxiv.org/abs/2405.20541v1
[112] G. Yang, D. Lo, R. D. Mullins, and Y. Zhao, "Dynamic Stashing Quantization for Efficient Transformer Training," *Findings of the Association for Computational Linguistics: EMNLP 2023*, pp. 7329–7336, Mar. 2023, doi: 10.18653/v1/2023.findings-emnlp.489.
[113] H. Zhang, G. Li, J. Li, Z. Zhang, Y. Zhu, and Z. Jin, "Fine-Tuning Pre-Trained Language Models Effectively by Optimizing Subnetworks Adaptively," *Adv Neural Inf Process Syst*, vol. 35, Nov. 2022, Accessed: Oct. 13, 2024. [Online]. Available: https://arxiv.org/abs/2211.01642v1